\pdfoutput=1
\documentclass[11pt]{article}

\usepackage[margin=1in]{geometry}

\usepackage[dvipsnames]{xcolor}

\usepackage{amsmath, amssymb}

\usepackage{array, booktabs, multirow}

\usepackage[ruled,vlined,linesnumbered]{algorithm2e}

\usepackage{tikz}
\usetikzlibrary{positioning, shapes.geometric, arrows.meta, fit, calc, backgrounds, decorations.pathreplacing}
\usepackage{pgfplots}
\pgfplotsset{compat=1.18}

\usepackage[numbers,sort&compress]{natbib}

\usepackage[colorlinks=true,linkcolor=blue!60!black,citecolor=blue!60!black,urlcolor=teal]{hyperref}
\hypersetup{
  pdftitle={Nous: An Attempt to Extract and Inject the Cognition Behind Prediction-Market Behavior},
  pdfauthor={Haowei Qian},
  pdfkeywords={cognitive diversity, prediction markets, LLM agents, behavioral extraction, epistemic monoculture, ensemble forecasting}
}

\usepackage{subcaption}
\usepackage{graphicx}

\usepackage{enumitem}
\usepackage{microtype}

\usepackage{titlesec}
\titleformat{\section}{\large\bfseries}{\thesection}{1em}{}
\titleformat{\subsection}{\normalsize\bfseries}{\thesubsection}{1em}{}

\usepackage{setspace}
\usepackage{fancyhdr}
\newcommand{\nous}{\textsc{Nous}}
\newcommand{\eg}{e.g.}

\providecommand{\checkmark}{\ensuremath{\surd}}

\title{\textbf{Nous: An Attempt to Extract and Inject\\the Cognition Behind Prediction-Market Behavior}}

\author{
  Haowei Qian \\
  Independent Researcher \\
  \texttt{qianhaowei@alu.fudan.edu.cn}
}

\date{}

\begin{document}

\maketitle

\begin{abstract}
As large language model (LLM) agents proliferate in prediction markets and collective decision-making, they risk a \emph{cognitive monoculture}: agents built on shared foundation models produce correlated forecasts, and recent measurement finds the errors of independently-developed frontier models correlated at $r \approx 0.77$. We ask whether human cognitive diversity can be recovered from behavior and transferred to LLM agents. \nous{} extracts a structured eight-dimension behavioral profile from real Polymarket trading activity and injects it into agents through prompts. Our central finding is a dissociation between the two halves of that pipeline. \textbf{Extraction works, partially}: across 100 wallets, 8 of 14 parameters are temporally stable (split-half ICC $\geq 0.5$, bootstrap CI lower bound $>0.3$; contrarian score reaches ICC $\approx 0.9$); wallets are identifiable from their profiles well above chance (top-1 retrieval $17$--$22\%$ vs.\ a $1\%$ random baseline); and two of four pre-specified dimensions rank-correlate with future realized profit out-of-sample, though the correlations do not survive behavioral-confound controls. \textbf{Prompt-level injection does not measurably transmit it}: on a semantic embedding metric, structured injection shows no significant advantage over a length-matched control on any model, and the diversity it induces neither reduces ensemble error correlation nor improves Brier score---a null that persists across exploratory checks on sampling temperature ($0.0$--$1.0$), a more-diverse synthetic profile population, and the model-uncertain question subset. Measuring the prompts themselves locates a compressing stage before the model: the structure-to-narrative translator emits semantically near-uniform prompts whose spread does not increase when profile spread does. We therefore position \nous{} as measuring the cognitive-monoculture problem and the limits of a prompt-level remedy, motivating deeper, below-the-prompt injection (parameter-efficient fine-tuning, activation steering). We release the evaluation code, extracted profiles, generated prompts, and model outputs at \url{https://github.com/WillChienT/nous-paper}; the extraction parameters fitted to live user data remain gray-box.
\end{abstract}

\vspace{0.5em}
\noindent\textbf{Keywords:} cognitive diversity, prediction markets, LLM agents, behavioral extraction, epistemic monoculture, ensemble forecasting

\section{Introduction}
\label{sec:intro}

\subsection{The 2026 Prediction Landscape}

The question of whether AI systems can match human forecasting accuracy has, in the past two years, moved from open conjecture to established empirical fact. \citet{halawi2024forecasting} demonstrated that a structured research-prediction pipeline---combining retrieval-augmented language models with multi-step reasoning---could approach the aggregate accuracy of human forecasting crowds on a diverse benchmark of real-world binary questions. That result reframed the conversation: not whether LLMs could forecast at all, but how close a principled system architecture could push aggregate performance toward the human crowd frontier.

The Mantic system extended this further in 2026, applying GRPO-style reinforcement learning with Brier score as the reward signal to a 120-billion-parameter open-source base model: the system's Metaculus baseline score rose from 38.6 to 45.8, and in the Metaculus Cups it placed 8th of roughly 550 entrants in Summer 2025 (the first AI system in the top ten of a large-scale forecasting tournament) and 4th in Fall 2025~\citep{Mantic2026Blog}. What the result demonstrates is the maturity of the approach: reinforcement learning on verifiable prediction questions can produce forecasting behavior that competes with professional human forecasters. Concurrent benchmark work maps where residual gaps remain---imprecise temporal alignment, over-conservative reading of ambiguous evidence, slow updating in the final hours before resolution~\citep{ProphetArena2026}; Section~\ref{sec:related:llmforecasting} reviews these lines in detail.

These lines together mark a meaningful inflection point. Recent systems have made single-agent forecasting accuracy a mature and rapidly advancing target; population-level error dependence, by contrast, remains comparatively underexplored. The question we take up is what happens when AI systems operate not as individual agents but as populations.

\subsection{Cognitive Homogeneity as the Next Structural Gap}

When many AI agents are deployed in the same information environment, accuracy at the individual level does not guarantee quality at the collective level. Agents built on the same foundation model produce highly correlated reasoning trajectories: when one agent misforecasts an event, agents trained on the same weights tend to reach the same wrong conclusion through the same structural failure. This correlation is not a coincidence of poor implementation but a consequence of shared pretraining on overlapping corpora with aligned fine-tuning objectives. A thousand GPT-based agents may represent only a handful of genuinely independent reasoning pathways. Recent measurement makes this concrete: across 568 resolved questions, the forecasting errors of three independently-developed frontier models (GPT-4o, Claude, Gemini) correlate at $r \approx 0.77$--$0.78$ \citep{oraclefingerprint2026}, leaving an ostensibly diverse ensemble with little genuine independence---a quantitative signature of epistemic monoculture. The pattern is reinforced by evidence that default LLM forecasters share systematic cognitive biases \citep{wang2025beyondbias} and that, left untuned, they behave markedly unlike the human traders whose diversity prediction markets were built to aggregate \citep{henning2025llmtraders}.

\citet{schoenegger2024silicon} documented this empirically in a multi-LLM forecasting ensemble. Drawing on GPT-4, Claude, Gemini, and other frontier models, the ensemble achieved accuracy improvements over any individual model---consistent with the classical result that diverse crowds outperform high-ability individuals under appropriate conditions~\citep{hong2004groups}. But the diversity that drove those improvements came from architectural differences between model families, not from differences in how agents attended to information, weighted evidence, or perceived risk. Within a single model family, agents remained highly correlated in their errors.

This matters acutely in prediction markets. \citet{surowiecki2004wisdom} established that collective intelligence rests on four conditions: diversity of opinion, independence of judgment, decentralization, and effective aggregation mechanisms. Markets satisfy the aggregation condition through price discovery. Human participation has historically approximated the first three through the natural variation in how different people attend to information, perceive risk, and update beliefs in response to new evidence. When AI systems displace or supplement human participants, all three enabling conditions degrade simultaneously: diversity narrows to a handful of foundation models trained on overlapping corpora; independence is undermined because shared weights produce correlated processing of the same inputs; decentralization becomes superficial when the cognitive origins of many agents trace back to a small number of training pipelines.

The missing piece is not more models but more \emph{cognitive structure} variation. The ensemble approach in \citet{schoenegger2024silicon} used different model families as a proxy for different reasoning styles; this cannot scale to the case where an operator wants to deploy many agents from a single high-quality model. The question we ask is whether structured cognitive heterogeneity can be induced through an injection mechanism rather than through architectural multiplicity---and whether the resulting diversity is meaningful enough to reduce the correlated failure modes that are the practical consequence of cognitive monoculture.

\subsection{Contributions}

We present \nous{}, a framework for extracting structured behavioral proxies from human prediction market activity and testing whether they can be injected into LLM agents to induce cognitive heterogeneity. Our contributions are organized around three falsifiable claims, each with a corresponding experimental test, and they form a dissociation: the source signal is partially recoverable, but the prompt-level bridge does not carry it into useful agent diversity.

\begin{enumerate}[leftmargin=*,itemsep=2pt]
    \item \textbf{Recoverability (the main result)}: A structured eight-dimension schema recovers a reliable \emph{subset} of behavioral-proxy dimensions from real prediction-market behavior. On $N=100$ Polymarket wallets, 8 of 14 parameters are temporally stable (split-half ICC $\geq 0.5$ with bootstrap CI lower bound $>0.3$), led by \texttt{independence.contrarian\_score} (ICC $\approx 0.9$); wallets are identifiable from their profile well above chance (top-1 retrieval $17$--$22\%$ vs.\ a $1\%$ random baseline); and two of four pre-specified dimensions predict realized trading profit \emph{out-of-sample}, once the same-source circularity between profile and PnL is removed (though the correlations do not survive behavioral-confound controls). Risk-perception parameters and clean individual-trajectory separability do not reach threshold---results we report rather than set aside (Section~\ref{sec:recoverability}).

    \item \textbf{Injectability (negative under the proper metric)}: Injected personas change agent outputs relative to a no-instruction control, but we cannot establish that the structured content transmits the profile \emph{beyond prompt length}. On a semantic embedding metric, no model shows a statistically significant advantage of the structured prompt over a length-matched filler; the favorable reading appears only on a lexical (trigram) metric and never reaches significance; and claims that the model conditions on the correct persona, or that the effect localizes to specific prompt content, do not survive (Section~\ref{sec:injectability}).

    \item \textbf{Usefulness (negative)}: The output diversity that injection does produce does \emph{not} improve collective forecasting. Cognitive injection raises ensemble output diversity slightly (pure-cognitive Jensen--Shannon divergence $+0.0035$, $p=0.010$, placebo-controlled for prompt length) but does not reduce ensemble error correlation or improve Brier score ($p \approx 0.95$), a null that persists across exploratory checks on sampling temperature $0.0$--$1.0$, a more-diverse synthetic population, and the model-uncertain question subset (Section~\ref{sec:usefulness}).
\end{enumerate}

Three features of the evidence locate the bottleneck. Structured injection shows no significant advantage over a length-matched control; a deliberately more-diverse profile population produces no more output divergence than the real one; and measuring the translator's outputs directly shows why---the persona prompts it emits are semantically near-uniform, and their spread does not increase when the spread of the input profiles does (Section~\ref{sec:usefulness:heterogeneity}). Spreading the inputs does not spread the prompts, let alone the outputs. This points to the injection \emph{channel}, the prompt and its structure-to-narrative translator, as the limiting factor, rather than the extracted signal or, on the evidence we have, the base model's capability. We therefore present \nous{} as measuring the limits of prompt-level cognitive injection and motivating deeper, below-the-prompt methods, not as a system that resolves cognitive monoculture.

The remainder of this paper is organized as follows. Section~\ref{sec:related} surveys related work across LLM forecasting, multi-agent ensembles, persona research, and behavioral economics. Section~\ref{sec:framework} defines the \nous{} Schema and the extraction-injection architecture. Sections~\ref{sec:recoverability}, \ref{sec:injectability}, and~\ref{sec:usefulness} present the three experiments in sequence. Section~\ref{sec:discussion} discusses implications and limitations. Section~\ref{sec:conclusion} concludes.

\section{Related Work}
\label{sec:related}

\subsection{LLM Forecasting: From Capability Question to Structural Question}
\label{sec:related:llmforecasting}

Research on LLM-based forecasting has moved quickly from asking whether language models can predict world events to characterizing the specific conditions under which they perform well or poorly.

\citet{halawi2024forecasting} established the first large-scale empirical baseline: a two-phase pipeline combining retrieval-augmented research with structured probabilistic reasoning could approach the aggregate accuracy of human forecasting crowds on a diverse benchmark of real-world binary questions. Their key finding was architectural rather than model-specific---the research phase, which gathered and synthesized relevant evidence before the prediction phase produced a calibrated estimate, accounted for most of the performance improvement over direct querying.

The Mantic system pushed this further in 2026 by adding reinforcement learning. Starting from a 120-billion-parameter open-source base model, the team applied GRPO-style policy gradient RL with Brier score as the reward signal, training on approximately ten thousand binary prediction questions across a two-phase architecture (a research phase followed by a mixture-model prediction phase). The baseline Metaculus score improved from 38.6 to 45.8---slightly above Gemini 3 Pro. In the Metaculus Cups, the system placed 8th of roughly 550 entrants in Summer 2025, the first AI system in the top ten of a large-scale forecasting competition, and 4th in Fall 2025~\citep{Mantic2026Blog}. The Mantic team's subsequent ensemble analysis found that model components with the highest Jensen-Shannon divergence from the ensemble's aggregate distribution were the least replaceable, suggesting that heterogeneity in prediction behavior---not just accuracy---determines marginal contribution to ensemble quality.

The Prophet Arena benchmark provides systematic measurement of where LLM forecasting still lags. Three gaps emerge consistently: models are imprecise on event timing when the resolution criterion involves exact dates; they are more conservative than human market participants in interpreting ambiguous source evidence; and they are slow to react in the final hours before question resolution, when human markets update aggressively on near-term signals~\citep{ProphetArena2026}. A growing benchmarking landscape situates these findings: \citet{predictionarena2026} evaluate LLM forecasters across question types, and \citet{polybench2026} assemble a large multimodal Polymarket benchmark (markets, order-book data, and news) of the kind our own evaluation set in Section~\ref{sec:usefulness:mantic} draws from. A complementary observation motivates our entire approach: \citet{henning2025llmtraders} find that untuned LLMs behave far more rationally than human traders and conclude that they should \emph{not} be used to mimic human behavioral data---which is precisely why an explicit injection mechanism, rather than a default agent, is needed to restore human-like cognitive variation.

\nous{} is not positioned as a competitor to these approaches. Mantic's optimization target is single-agent prediction accuracy; its training objective pushes a single system toward the behavior of expert human forecasters. Our target is different: we ask how a population of agents---potentially sharing the same base model---can maintain meaningful cognitive differences from one another, so that their errors are not highly correlated. These are orthogonal objectives. A Mantic-accurate model that exhibits cognitive homogeneity across its instances will still fail the collective intelligence test described in Section~\ref{sec:intro}; a population of \nous{}-diverse agents that are individually less accurate may still produce better ensemble forecasts if their errors are independent. The empirical question of which matters more is left to Section~\ref{sec:usefulness}.

\subsection{Multi-Agent Ensembles and Cognitive Diversity}
\label{sec:related:ensembles}

The theoretical case for diverse agent populations follows directly from the crowd wisdom literature. \citet{hong2004groups} showed formally that, under specific conditions, groups of diverse problem-solvers can outperform groups of high-ability but homogeneous solvers. \citet{surowiecki2004wisdom} identified the necessary conditions: diversity of opinion, independence of judgment, decentralization, and effective aggregation. The challenge for AI-populated systems is that architectural uniformity violates the first two conditions systematically.

\citet{schoenegger2024silicon} demonstrated both the promise and the limitation of current ensemble approaches. Aggregating forecasts from multiple frontier models---GPT-4, Claude, Gemini, and others---produced accuracy improvements consistent with the crowd wisdom prediction. But the diversity in that ensemble came from training differences between model families, not from differences in how agents process and weight information. The gap \nous{} addresses is the case where architectural diversity is unavailable or impractical, and cognitive diversity must instead come from the structure of how agents are configured.

Multi-agent debate methods offer a related but distinct approach. \citet{du2023debate} showed that having multiple agents argue for different positions before reaching a consensus improves factual accuracy and reasoning consistency. However, debate methods introduce diversity through stochastic sampling and role assignment rather than through structured cognitive parameters. The diversity is procedural rather than dispositional: two debate agents drawing from the same model weights remain correlated in their priors, even when assigned opposing positions.

A more directly relevant methodological comparison is the use of Jensen-Shannon divergence to quantify how replaceable individual ensemble members are. The Mantic ensemble analysis found that models with high JS divergence from the aggregate provided disproportionate value. \nous{} aims to achieve this kind of irreplaceability not through architectural differences but through structured injection of distinct cognitive profiles---profiles grounded in behavioral observations from genuinely different human forecasters. That this irreplaceability is scarce among current models is now an empirical fact: \citet{oraclefingerprint2026} measure forecasting-error correlations of $r \approx 0.77$--$0.78$ across three independently-developed frontier models on 568 resolved questions---little effective diversity among systems built by different organizations, the epistemic monoculture our injection mechanism is designed to counteract.

The work closest to \nous{} in surface form is the use of multiple LLM personas on prediction markets. \citet{polyswarm2026} deploy roughly fifty LLM personas on Polymarket binary markets and aggregate their forecasts, reporting Brier score and trading profit. Despite the apparent similarity, the two systems differ on four axes that define our contribution. First, persona \emph{source}: their personas are hand-designed by the authors as free-form character sketches, whereas \nous{} profiles are extracted empirically from the on-chain behavior of real traders. Second, \emph{schema}: they impose no structured dimensional representation, whereas \nous{} profiles are an eight-dimension continuous parameterization. Third, \emph{evaluation}: they measure accuracy and profit but not ensemble heterogeneity, whereas our central metrics are JSD, pairwise error correlation, and ensemble Brier. Fourth, \emph{baseline}: they include no identical-agent control, so the contribution of persona diversity cannot be separated from base-agent capability, which is exactly the comparison our placebo-controlled heterogeneity experiment (Section~\ref{sec:usefulness:heterogeneity}) is built around.

A second cluster of recent work builds heterogeneous-persona ensembles for forecasting but along different axes. \citet{promptingpolicy2025} assemble a large ensemble of demographic personas for macroeconomic point forecasts and study forecast dispersion; their personas are sampled generic demographic profiles rather than behavioral reconstructions, and their target is a continuous macro quantity rather than market-resolved binary questions. \citet{twinmarket2025} reconstruct LLM agents from real trader data---sharing our ``behavioral reconstruction'' premise---but their goal is agent-based market \emph{simulation} of price and social dynamics, not a structured cognitive schema evaluated for ensemble forecast diversity. Outside forecasting entirely, \citet{multiperspectivity2026} report that practitioner personas in a narrative-interpretation task are individually weaker but make less correlated errors, so that majority voting benefits more---a cross-domain corroboration of the persona-heterogeneity-to-decorrelation principle that underlies our usefulness hypothesis.

\subsection{Persona and Prompting Research}
\label{sec:related:persona}

A substantial body of work has explored whether LLM agents can be configured to represent distinct cognitive styles through prompting.

\citet{park2023generative} introduced generative agents with memory, reflection, and planning, demonstrating that rich behavioral simulation is achievable with language models. However, persona specification in that work was manual and descriptive---character sketches rather than structured cognitive parameters. There was no pipeline for grounding an agent's behavioral tendencies in observations of real human behavior.

\citet{shanahan2023role} analyzed role-playing with LLMs systematically, showing that persona prompting produces surface-level behavioral changes but that the underlying model behavior remains constrained by training. \citet{salewski2024context} provided a more precise characterization: in-context impersonation reveals models' strengths and biases, but does not give access to the full range of human cognitive variation. A model instructed to ``think like a risk-seeking investor'' will produce risk-seeking-sounding output, but its underlying processing is still governed by the mean-field representation learned during training.

A growing body of more recent work examines cognitive architecture specifically. \citet{bao2025conflictaware} address logical reasoning under contradictory evidence through a dual-process architecture. \citet{yang2026thinkfastslow} introduce CogRouter, which dynamically adapts cognitive depth at each reasoning step within a single agent. \citet{zhu2024eliciting} demonstrate that LLMs encode recoverable probabilistic priors that qualitatively align with human priors.

\citet{wang2025beyondbias} are the closest in stated aim: their multi-cognition agentic framework assigns diverse cognitive perspectives to reduce forecasting bias, and their finding that several LLMs share systematic in-group, premise-induced, and confirmation biases is direct evidence for the monoculture this paper addresses. The mechanism, however, is distinct from ours. Their ``cognition'' is the stance of an event \emph{stakeholder}---the model role-plays a country, organization, or individual relevant to the question, with the roster of stakeholders mined from question topics. \nous{} ``cognition'' is instead a trader's structured parameter vector, reconstructed from on-chain behavior rather than generated from the question, injected through a structure-to-narrative translator rather than a role-play instruction, and evaluated for ensemble diversity rather than single-system accuracy.

A separate line grounds personas in real human data, as we do, but with different scope and validation. \citet{kitadai2025biasadjusted} inject behavioral-economics parameters estimated from ultimatum-game experiments into agents, validating against responder-side behavioral alignment rather than market forecasting. \citet{designingpersonalities2026} reconstruct Big Five personality profiles from human survey data and inject them into agents, validating persona fidelity rather than any downstream forecasting or ensemble task.

These works treat cognitive characteristics as universal architectural features, population-level biases, or general personality traits. What none provides is the specific combination \nous{} targets: a structured trading-cognition schema, reconstructed from individual on-chain behavior, and evaluated by whether the resulting ensemble's errors decorrelate. \nous{} is orthogonal to this literature: rather than improving how a single agent reasons, we use structured profiles from real human behavior to ensure that different agents reason differently.

The gap we address is the step from ``surface prompting changes observed outputs'' to ``does the underlying reasoning structure change?'' Our Injectability experiment (Section~\ref{sec:injectability}) is designed to test this directly, using length-controlled and persona-mismatch baselines to isolate the contribution of cognitive structure from prompt length effects.

\subsection{Behavioral Economics Anchors for the Eight Dimensions}
\label{sec:related:behavioralecon}

The \nous{} Schema draws its theoretical grounding from several converging lines of research in behavioral economics and judgment under uncertainty.

\citet{kahneman1979prospect} established that human risk perception follows an asymmetric value function with individually varying curvature parameters ($\alpha$, $\beta$) and loss aversion coefficient ($\lambda$). This provides the mathematical basis for the risk perception dimension: different individuals have genuinely different Prospect Theory parameters, not just different surface risk attitudes. The empirical literature on parameter estimation~\citep{tversky1992advances} establishes population distributions for these parameters ($\alpha = \beta = 0.88$, $\lambda = 2.25$ as reference means), which \nous{} uses as cold-start priors when behavioral data is sparse.

\citet{tetlock2005expert} identified the cognitive traits that distinguish consistently accurate forecasters: actively open-minded thinking, granular probabilistic reasoning, and fox-like integration of diverse information sources, as opposed to the hedgehog-like reliance on a single explanatory framework. \citet{tetlock2015superforecasting} extended this work, documenting that superforecasters update beliefs frequently and in proportion to new evidence. These findings directly inform the belief update inertia and cognitive style (hedgehog-fox) dimensions of the schema.

\citet{surowiecki2004wisdom} provides the collective intelligence framing that motivates the entire project: individual cognitive variation is not noise to be averaged out but the substrate that makes collective judgment possible. The diversity condition in Surowiecki's framework requires that different participants process information through genuinely different cognitive lenses---which is precisely what the eight-dimension schema is designed to capture and inject.

\citet{galef2021scout} provides a modern behavioral formulation of the hedgehog-fox distinction that proved useful in operationalizing the cognitive style dimension. The scout mindset---seeking accurate information even when it contradicts prior beliefs---maps directly to the fox-type parameters in the schema and provides a behavioral anchor for the independent thinking dimension.

References to these anchors are expanded in the context of specific dimensions in Section~\ref{sec:framework}; the purpose of collecting them here is to establish that the schema dimensions are not arbitrary but each traces to a line of empirical research on human cognitive heterogeneity.

\section{Framework: Structured Behavioral Proxies for Agent Heterogeneity}
\label{sec:framework}

\subsection{The Eight-Dimension Schema}
\label{sec:framework:schema}

\paragraph{Formal definition.}
We define a \emph{structured behavioral proxy} as a pre-reasoning information processing function inferred from observed behavioral data, which shapes how inputs are filtered, weighted, and framed before deliberative reasoning begins. Formally, let $x$ denote raw information and $r(\cdot)$ a reasoning function. Standard LLM processing computes $r(x)$ directly. We propose interposing a behavioral proxy $\phi$:
\begin{equation}
    \text{output} = r(\phi(x))
\end{equation}
where $\phi: \mathcal{X} \to \mathcal{X}$ transforms the information space by adjusting salience, risk framing, temporal weighting, and domain attention. Behavioral proxies are distinct from knowledge (factual information in memory), skills (procedural capabilities), personality (values and social preferences), and memory (episodic interaction records). They determine \emph{how} information is preprocessed, not \emph{what} is known or \emph{how} tasks are executed.

We use the term ``behavioral proxy'' rather than the stronger ``cognitive prior'' deliberately. What the extraction pipeline recovers from trading behavior is a structured set of observable parameters---bet size distributions, holding durations, domain frequencies, adjustment patterns. These are proxies for underlying cognitive structure, not direct measurements of it. The gap between behavioral observable and cognitive reality is a limitation we discuss explicitly in Section~\ref{sec:framework:claims}.

\paragraph{Why the proxy layer is not spontaneously produced.}
The absence of individual cognitive variation in LLM outputs is not empirical accident but structural consequence. The standard language modeling loss,
\begin{equation}
    \min_\theta \; \mathbb{E}_{(x,y) \sim \mathcal{D}} \left[ -\log P_\theta(y \mid x) \right],
\end{equation}
computes an expectation over billions of human-generated texts---mathematically equivalent to learning a mean-field approximation of human cognition. Individual cognitive differences are treated as noise in the aggregate. Alignment training further compresses this distribution: RLHF~\citep{ouyang2022training} and DPO~\citep{rafailov2023direct} optimize toward majority-vote human preferences, systematically downweighting minority cognitive patterns that contribute most to epistemic diversity.

The consequence is that without external injection, a prompted agent converges to a mode-of-training-data reasoning trajectory. This is what we call the \emph{structural default}: not that the model is incapable of deviating, but that deviation requires explicit external input. We return to this reframing in Section~\ref{sec:framework:claims}, where it resolves a tension present in an earlier version of this work.

\paragraph{Grey-box design.}
We adopt a gray-box approach: the structure of a behavioral proxy is visible (eight named dimensions in three layers), but the precise parameters are not publicly disclosed. This balances three requirements. Fully opaque proxies prevent diversity measurement and trust calibration by downstream consumers. Fully transparent proxies invite gaming---adversaries could construct inputs that exploit known biases, and exact transparency enables replication, reintroducing the monoculture problem. The gray-box solution exposes structure (``this agent has high loss aversion and focuses on politics'') while keeping exact parameterization ($\lambda = 2.73$) internal.

\paragraph{Core-Shell-Membrane architecture.}
Inspired by stability gradients in human cognition~\citep{kahneman2011thinking, klein1998sources}, we organize the \nous{} Schema in three concentric layers with decreasing stability (Figure~\ref{fig:csm}):

\begin{figure}[t]
    \centering
    \resizebox{0.7\textwidth}{!}{\begin{tikzpicture}[
    font=\sffamily\small
]

\fill[orange!15] (0,0) circle (3.8cm);
\draw[orange!60, thick, dashed] (0,0) circle (3.8cm);

\fill[teal!20] (0,0) circle (2.7cm);
\draw[teal!60, thick] (0,0) circle (2.7cm);

\fill[blue!25] (0,0) circle (1.5cm);
\draw[blue!70, thick] (0,0) circle (1.5cm);

\node[font=\sffamily\footnotesize\bfseries, text=blue!80] at (0,0.5) {Core};
\node[font=\sffamily\tiny, text=blue!60, align=center] at (0,-0.2) {Risk Perception\\Time Scale\\Cognitive Style};

\node[font=\sffamily\footnotesize\bfseries, text=teal!80] at (0,2.1) {Shell};
\node[font=\sffamily\tiny, text=teal!60, align=center] at (2.0,0) {Attention\\Allocation};
\node[font=\sffamily\tiny, text=teal!60, align=center] at (-2.0,0) {Belief Update\\Inertia};
\node[font=\sffamily\tiny, text=teal!60, align=center] at (0,-2.0) {Domain\\Confidence};

\node[font=\sffamily\footnotesize\bfseries, text=orange!70] at (0,3.3) {Membrane};
\node[font=\sffamily\tiny, text=orange!60] at (3.0,1.8) {Independence};
\node[font=\sffamily\tiny, text=orange!60] at (-3.0,1.8) {Loss Response};

\draw[-{Stealth[length=2.5mm]}, thick, blue!60]
    (5.2, -2.5) -- (5.2, 2.5)
    node[midway, right=3pt, font=\sffamily\tiny, text=blue!60, align=left] {Stability};

\draw[-{Stealth[length=2.5mm]}, thick, orange!60]
    (6.2, 2.5) -- (6.2, -2.5)
    node[midway, right=3pt, font=\sffamily\tiny, text=orange!60, align=left] {Mutability};

\node[font=\sffamily\tiny, text=gray, align=center] at (-5.0, 1.0)
    {High reactivity\\Fast adaptation};
\node[font=\sffamily\tiny, text=gray, align=center] at (-5.0, -0.5)
    {Moderate inertia\\Slow evolution};
\node[font=\sffamily\tiny, text=gray, align=center] at (-5.0, -2.0)
    {Near-immutable\\Identity anchor};

\draw[gray!40, thin, ->] (-3.9, 1.0) -- (-3.8, 1.0);
\draw[gray!40, thin, ->] (-3.9, -0.5) -- (-2.7, -0.5);
\draw[gray!40, thin, ->] (-3.9, -2.0) -- (-1.5, -1.0);

\end{tikzpicture}}
    \caption{Core-Shell-Membrane architecture of the \nous{} Schema. Inner layers are more stable; outer layers are more reactive. Arrows indicate the stability--mutability gradient.}
    \label{fig:csm}
\end{figure}
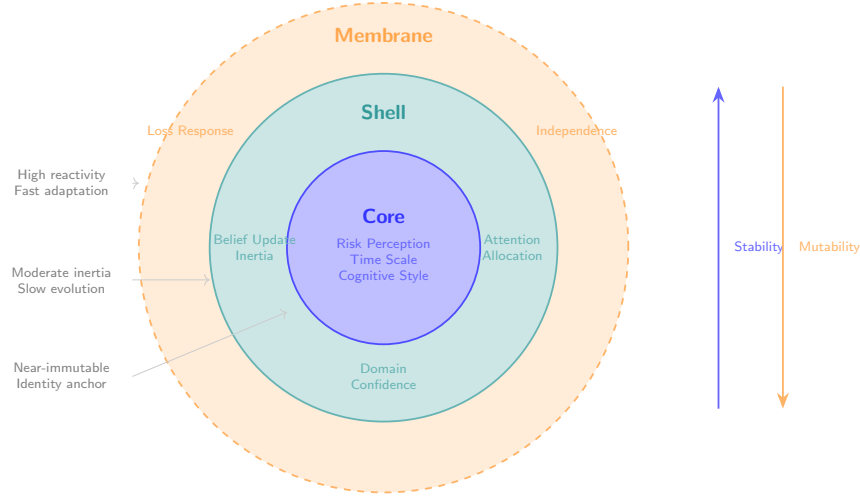

\begin{itemize}[leftmargin=*,itemsep=2pt]
    \item \textbf{Core} (near-immutable): Deep behavioral parameters that change rarely---risk perception, temporal orientation, and cognitive style. These are analogous to personality traits shaped by culture and formative experience; in the behavioral data, they manifest in consistent patterns across many independent trading decisions.
    \item \textbf{Shell} (slowly evolving): Domain-specialized parameters that adapt with sustained experience---attention allocation, belief update behavior, and domain-specific calibration. Requires many observations to shift and may change over months with sustained engagement in a new domain.
    \item \textbf{Membrane} (reactive): Real-time behavioral signals---independence from crowd and loss response patterns. These fluctuate with recent experience but do not permanently alter the underlying profile.
\end{itemize}

The design principle is that the deeper the layer, the higher the modification threshold and the slower the rate of change~\citep{tversky1992advances}. A human can change their mood in hours (membrane), revise a professional judgment over months (shell), but rarely alters their foundational risk attitudes (core). The schema mirrors this stability gradient.

\paragraph{Eight dimensions.}
Table~\ref{tab:dimensions} specifies the eight dimensions, their theoretical grounding, and the minimum behavioral events required for extraction above population-prior confidence. Two dimensions---\texttt{cognitive\_style} and \texttt{independence\_index}---carry an additional caveat: the behavioral proxies measured from trading data (decision latency ratios, category concentration, crowd-divergence frequency) are more distal from the underlying cognitive construct than dimensions like risk perception, where a principled model (Prospect Theory) connects observable bets directly to parameters. Downstream users should interpret cognitive style and independence values as behavioral tendencies rather than direct cognitive measurements.

\begin{table}[t]
\centering
\caption{The eight dimensions of the \nous{} Schema. Each dimension specifies parameter ranges, theoretical grounding, and minimum events required for extraction above population-prior confidence. \dag~Dimensions marked with \dag{} are measured from observable trading tendencies; the connection to underlying cognitive structure is more distal than for Core dimensions with principled behavioral models.}
\label{tab:dimensions}
\small
\begin{tabular}{@{}llllll@{}}
\toprule
\textbf{Dimension} & \textbf{Layer} & \textbf{Parameters} & \textbf{Range} & \textbf{Theory} & \textbf{Min.} \\
\midrule
Risk Perception & Core & $\alpha$, $\beta$, $\lambda$ & [0,1], [0,1], [1,5] & Prospect Theory~\citep{kahneman1979prospect} & 50 \\
Time Scale Pref. & Core & horizon\_bias, patience & [-1,1], [0,1] & Tetlock~\citep{tetlock2015superforecasting} & 20 \\
Cognitive Style\dag & Core & analytical\_ratio, HF & [0,1], [-1,1] & System 1/2~\citep{kahneman2011thinking} & 30 \\
\midrule
Attention Alloc. & Shell & domain\_weights, blind\_spots & $\Delta^K$, list & Klein RPD~\citep{klein1998sources} & 20 \\
Belief Update & Shell & update\_rate, conf\_bias & [0,1], [0,1] & Griffiths~\citep{griffiths2008bayesian} & 30 \\
Domain Conf. & Shell & per-domain accuracy, freq. & [0,1], $\mathbb{N}$ & Klein--Kahneman~\citep{kahneman2009conditions} & 15 \\
\midrule
Independence\dag & Membrane & contrarian, crowd\_sens. & [-1,1], [0,1] & Tetlock~\citep{tetlock2005expert} & 30 \\
Loss Response & Membrane & sunk\_cost, tilt, resilience & [0,1], [0,1], $\mathbb{R}^+$ & Damasio~\citep{damasio1994descartes} & 25 \\
\bottomrule
\end{tabular}
\end{table}

\paragraph{Confidence metadata and graceful degradation.}
Each dimension carries confidence metadata: a composite score derived from data sufficiency (sigmoid function of event count relative to threshold), temporal stability (coefficient of variation across time windows), and internal consistency. When confidence falls below a threshold ($\tau = 0.2$), the dimension falls back to population priors from the behavioral economics literature (\eg{} $\alpha = \beta = 0.88$, $\lambda = 2.25$ from \citet{kahneman1979prospect}). This graceful degradation ensures that a \nous{} Instance is always well-defined, even for users with sparse behavioral history. Importantly, the Recoverability experiment (Section~\ref{sec:recoverability}) reveals that confidence thresholds interact with dimension reliability: some dimensions that appear to have sufficient data volume still exhibit low split-half ICC, suggesting that event count alone is an insufficient proxy for extraction quality.

\subsection{Behavioral Extraction Pipeline}
\label{sec:framework:extraction}

The extraction pipeline converts a stream of behavioral events from prediction market interactions into a structured \nous{} Instance. At a high level, the pipeline operates in three stages.

The first stage ingests a user's behavioral history $H = \{(a_t, c_t, o_t)\}_{t=1}^T$, where $a_t$ is an action (order, adjustment, view, hesitation), $c_t$ is the decision context (market price, category, recent PnL), and $o_t$ is the observed outcome. We define ten event types that capture meaningful behavioral signals: buy/sell orders, position closes with PnL, position adjustments, market page views, abandoned orders, canceled limit orders, comments, feed impressions with click-through, portfolio checks, and resolution events with Brier scores.

The second stage computes typed feature vectors per user. A \texttt{UserFeatures} object aggregates events by type, providing structured access to orders (with decision latency, order type, probability, size), position closes (with PnL and hold duration), adjustments (with price change context and direction), and interaction signals (views, hesitations, feed impressions). This intermediate representation decouples event ingestion from dimension-specific extraction.

The third stage applies eight stateless extractors to the feature vector, each returning dimension values plus confidence metadata. For example, the risk perception extractor fits the Prospect Theory value function to observed bet sizes as a function of market probability; the attention allocation extractor computes exponentially-decay-weighted domain frequencies. When an extractor's output confidence falls below $\tau$, the corresponding dimension reverts to population priors.

The full algorithmic specification of each extractor---including the Bayesian inverse-problem framing and the MAP estimation approximations currently in use---is presented in Section~\ref{sec:recoverability}. We note here only that the pipeline is designed as a point-estimate approximation to a principled Bayesian framework, with known limitations treated as future work.

\subsection{Structure-to-Narrative Translator}
\label{sec:framework:translator}

A parametric \nous{} Instance must be converted into natural-language instructions before it can be injected into an LLM agent's system prompt. This conversion is performed by the structure-to-narrative translator, which exists in two versions.

The first-generation translator (v1) converts each dimension into qualitative narrative through discrete threshold-based branching: continuous parameters are quantized into 2--3 discrete buckets, each mapped to a fixed descriptive phrase. This approach is simple and interpretable but discards the fine-grained individual variation that the extraction pipeline captures.

The second-generation translator (v2) eliminates discrete bucketing in favor of five design principles: continuous parameter embedding (exact numerical values presented alongside population-relative z-scores), behavioral inference chains (each parameter value triggers a chain of behavioral consequences), cross-dimension interaction effects (explicitly described emergent behaviors from parameter combinations), contrastive anchoring (the prompt opens with the three largest deviations from population mean), and behavioral exemplars (concrete past-decision vignettes generated from the parameters). In multi-model experiments the v2 translator increased lexical (trigram) diversity over v1, but Section~\ref{sec:injectability} shows this gain is concentrated in a single model and does not extend to semantic or structured-output metrics; we no longer present it as matching handwritten baselines.

The detailed translator design, ablation results, and length-controlled baseline comparisons are presented in Section~\ref{sec:injectability}. We note here that the translator is the component most directly implicated in the prompt-length confound: the v2 translator produces substantially longer prompts than v1 or handwritten baselines, and disentangling the contribution of structural content from prompt length effects requires the length-controlled ablation that Section~\ref{sec:injectability} reports. A second, sharper implication emerges later: measured in embedding space, the translator's outputs are nearly uniform across very different profiles, identifying it as a compressing stage of the injection channel (Section~\ref{sec:usefulness:heterogeneity}).

\subsection{Claim Tightening: From Cognitive Priors to Behavioral Proxies}
\label{sec:framework:claims}

An earlier version of this work used the phrase ``cognitive priors'' to describe what the extraction pipeline recovers from prediction market behavior. This framing overclaims in two related ways, and the present version corrects both.

The first overclaim is the implied certainty about what is being measured. Bet sizes, holding durations, domain frequencies, and adjustment patterns are observable behavioral features. They correlate with underlying cognitive structure---a trader with high loss aversion tends to hold losing positions longer, and a trader with strong domain specialization concentrates bets in familiar categories---but the connection is mediated by confounding factors: market liquidity constraints, position limits, execution strategies, and external time commitments all influence trading behavior independently of cognitive structure. The Recoverability experiment (Section~\ref{sec:recoverability}) finds exactly this: the dimensions with the strongest behavioral-cognitive connection (independence, attention allocation) show high split-half ICC, while risk perception's curvature parameters---which depend on precise bet-size-to-probability mappings that are easily confounded by market microstructure---do not pass the reliability threshold. We adopt ``structured behavioral proxy'' as a more accurate description: structured because the schema provides a principled mapping from behavior to parameters; behavioral because what is directly measured is action, not cognition; and proxy because the connection to underlying cognitive structure is inferred, not observed.

The second overclaim is the description of cognitive diversity as a ``structural blind spot'' of LLM training, which implies LLMs are categorically incapable of exhibiting individual variation. The Injectability experiment contradicts this directly: on the models where our semantic metric discriminates, structured injection changes outputs relative to a no-instruction control (Section~\ref{sec:injectability}), which means the model is \emph{capable} of producing cognitively diverse outputs when given appropriate instructions---even though, as that section shows, we cannot attribute the change to structured content beyond prompt length. The blind-spot framing creates an internal contradiction that external review of the earlier version correctly identified.

The more accurate description is \emph{structural default}. Without external cognitive proxies, a prompted LLM converges to the mode-of-training-data reasoning trajectory---the mean-field behavior that the training objective selected. This default is the output of an averaging process that treats individual cognitive variation as noise. The default is not immutable: appropriate prompting can steer the model substantially away from it, as the injection experiments demonstrate. What is missing is a principled, automated method for steering consistently toward specific, grounded, extractable cognitive positions---and that is precisely what the \nous{} framework provides.

This reframing is not a concession. The motivation for \nous{} does not depend on claiming that LLMs are categorically unable to vary. It depends on claiming that they do not vary spontaneously in the structured, individually-grounded way that human participants in prediction markets vary---and that restoring this variation requires an explicit architectural mechanism. The structural default framing captures this motivation accurately without the overclaim.

To be precise about what this paper claims and does not claim, we state the three contributions in terms of their corresponding experiments:

The Recoverability claim is that a structured 8-dimension behavioral schema extracts stable, externally-valid proxies from real human prediction-market activity: stable by split-half ICC, structured at the population level under a valid permutation null, and predictive of realized profit on pre-specified dimensions. It is a claim about the extraction pipeline's reliability on specific dimensions and about population-level structure---not about clean individual-trajectory separability, which is fragile at our sample size, and not about recovering true cognitive structure.

The Injectability claim is negative: injected personas change agent outputs relative to a no-instruction control, but on a semantic embedding metric we cannot show that the structured content shifts outputs beyond a length-matched baseline on any model. It is a claim about the limits of prompt-level transfer, not a demonstration that structured content penetrates.

The Usefulness claim is bounded: cognitive injection produces a measurable but small increase in ensemble output diversity that does \emph{not}, at the magnitude prompt-level injection achieves, translate into improved ensemble Brier or reduced error correlation. The Heterogeneity Validation experiment of Section~\ref{sec:usefulness} establishes this limit rather than the collective-accuracy gain the framework ultimately seeks.

\section{Recoverability: Extracting Stable Behavioral Proxies from On-Chain Activity}
\label{sec:recoverability}

The first of our three claims is the most basic: that the eight-dimension schema, applied to real human prediction-market behavior, recovers \emph{stable} proxies rather than noise. This addresses the central validation gap of prior work on this framework: earlier validation was round-trip self-consistency on synthetic data, never a test against behavior the extractor did not itself generate. We test recoverability on the public on-chain trading history of Polymarket wallets, where the behavior is generated by real traders with no knowledge of our schema.

We decompose recoverability into three questions, each with a pre-specified test. Test~A asks whether a dimension is \emph{temporally stable}: does an estimate from the first half of a trader's history agree with an estimate from the second half? Test~B asks whether profiles are \emph{individually separable}: do distinct traders occupy distinguishable regions of the parameter space, beyond what a structure-free null produces? Test~C asks whether the recovered dimensions have \emph{predictive validity}: do they relate to an external behavioral outcome (realized profit) in the direction that the underlying theory predicts? The three tests answer to different standards of evidence, and they survive to different degrees, as we report below.

\subsection{Data and Cohort}
\label{sec:recoverability:data}

We collect the on-chain trade history of $N=100$ Polymarket wallets. Candidate wallets are drawn from the CLOB trade logs and filtered to remove automated market makers and arbitrage bots, which exhibit signatures (sub-second inter-trade latencies, near-symmetric two-sided quoting, thousands of trades per market) that would contaminate cognitive-trait estimation. The retained cohort is 100 wallets with at least five closed positions; in practice the selection skews to high-activity traders (a median of roughly two hundred closed positions per wallet), so the floor rarely binds. Effective sample sizes differ by analysis, and we report each: $N=100$ for the trade-index ICC (Test~A) and the predictive-validity tests (Test~C), and $N=84$ for the position-aware ICC, where wallets with too few closes in a time-ordered half are dropped. Trade events are reconstructed into positions using \texttt{position\_reconstructor.py}, which---following the resolution-payout correction described in Section~\ref{sec:recoverability:testc}---accounts both for positions closed by an offsetting trade in an open market and for positions held through to market resolution.

\subsection{On-Chain Adapter and Dimension Degradation}
\label{sec:recoverability:adapter}

The extraction pipeline of Section~\ref{sec:framework:extraction} was specified for a richer event stream than public on-chain data provides. The full schema consumes ten event types, including page views, abandoned orders, hesitation signals, and feed impressions---telemetry available to a first-party application but absent from the blockchain record, which exposes only executed trades, position changes, and resolution outcomes. We therefore route extraction through an on-chain adapter that maps the available events onto the extractors and flags every dimension whose inputs are only partially observable.

We refer to this flag as the \emph{degradation map}. A dimension is labeled \texttt{full-signal} when the on-chain record supplies the events its extractor was designed for, and \texttt{degraded} when the extractor must fall back to a sparser observable. Cognitive style (analytical ratio, hedgehog--fox index), loss response (sunk-cost sensitivity, tilt factor), and attention-allocation entropy are degraded under on-chain extraction, because the latency, hesitation, and cross-domain browsing signals they were designed around are not recorded on-chain. The degradation label is a statement about input completeness, not about the reliability of the resulting estimate; as the next section shows, some degraded-input dimensions are nonetheless highly reliable, while some full-signal dimensions are not.

\subsection{Test A: Temporal Stability (Split-Half ICC)}
\label{sec:recoverability:testa}

For each wallet we sort closed positions by trade index, split them into a first and second half, and extract an independent eight-dimension profile from each half. For every continuous parameter we then compute the intraclass correlation coefficient ICC(2,1) across the cohort, treating the two halves as repeated measurements of the same underlying trait. ICC(2,1) is conservative here: it penalizes both rank disagreement and absolute-level shifts between the halves. We adopt the conventional reliability thresholds ICC $\geq 0.5$ (moderate) and ICC $\geq 0.65$ (substantial).

Table~\ref{tab:testa} reports the result for all fourteen parameters, with the current reconstructor on $N=100$ wallets. Nine of the fourteen parameters clear the moderate threshold and six clear the substantial threshold; of the nine, eight retain a bootstrap 95\% CI lower bound above $0.3$ under the position-aware split (Table~\ref{tab:testa}), and we treat those eight as the robust set. The most reliable parameter is \texttt{independence\_index.contrarian\_score} (ICC $= 0.90$), followed by a cluster around ICC $\approx 0.75$: \texttt{attention\_allocation.entropy} (0.76), \texttt{independence\_index.crowd\_sensitivity} (0.76), \texttt{time\_scale\_preference.horizon\_bias} (0.75), and \texttt{cognitive\_style.analytical\_ratio} (0.73), with \texttt{belief\_update\_inertia.update\_rate} (0.68) completing the substantial tier.

\begin{table}[t]
\centering
\caption{Test~A split-half reliability across $N=100$ Polymarket wallets. ICC(2,1) is reported for both the trade-index split (ICC$_t$) and the position-aware split sorted by close time (ICC$_p$, $N=84$), with a bootstrap 95\% CI on the position-aware estimate and the rate at which the parameter fell back to a population prior. The \emph{Signal} column is the on-chain degradation label (Section~\ref{sec:recoverability:adapter}). Parameters are ordered by ICC$_t$; rules mark the substantial ($0.65$) and moderate ($0.50$) thresholds. The \checkmark{} column marks the robust set: the eight parameters with ICC$_t \geq 0.5$ and a position-aware CI lower bound $>0.30$ (for patience index the unrounded lower bound is $0.304$). No family-wise correction is applied, as the fourteen parameters are fixed by the schema rather than selected post hoc.}
\label{tab:testa}
\small
\begin{tabular}{@{}lcccclc@{}}
\toprule
\textbf{Parameter} & \textbf{ICC$_t$} & \textbf{ICC$_p$} & \textbf{95\% CI} & \textbf{Fallback} & \textbf{Signal} & \textbf{Robust} \\
\midrule
contrarian\_score & 0.90 & 0.90 & [0.84, 0.95] & 0\% & full & \checkmark \\
entropy & 0.76 & 0.76 & [0.62, 0.86] & 0\% & degraded & \checkmark \\
crowd\_sensitivity & 0.76 & 0.70 & [0.51, 0.83] & 0\% & full & \checkmark \\
horizon\_bias & 0.75 & 0.63 & [0.38, 0.82] & 0\% & full & \checkmark \\
analytical\_ratio & 0.73 & 0.79 & [0.68, 0.87] & 0\% & degraded & \checkmark \\
update\_rate & 0.68 & 0.62 & [0.44, 0.75] & 8\% & full & \checkmark \\
\midrule
patience\_index & 0.62 & 0.59 & [0.30, 0.81] & 0\% & full & \checkmark \\
hedgehog\_fox\_index & 0.59 & 0.54 & [0.28, 0.77] & 0\% & degraded & \\
$\lambda$ (loss aversion) & 0.57 & 0.61 & [0.46, 0.75] & 0\% & full & \checkmark \\
\midrule
$\alpha$ (gain curvature) & 0.44 & 0.43 & [0.15, 0.65] & 0\% & full & \\
$\beta$ (loss curvature) & 0.39 & 0.49 & [0.27, 0.67] & 0\% & full & \\
confirmation\_bias & 0.29 & 0.40 & [0.18, 0.61] & 8\% & full & \\
sunk\_cost\_sensitivity & 0.28 & 0.07 & [$-0.17$, 0.31] & 0\% & degraded & \\
tilt\_factor & 0.03 & 0.00 & [$-0.20$, 0.29] & 0\% & degraded & \\
\bottomrule
\end{tabular}
\end{table}

Two features of this result are worth drawing out, because both run counter to the priors stated in our prior design specification. First, the dimensions that recover most reliably are not the ones with the most principled behavioral model. Risk perception is grounded in Prospect Theory, with a closed-form value function relating bet size to market probability, yet its three parameters are the least reliable of the Core dimensions: $\lambda$ reaches only 0.57 and the curvature parameters $\alpha,\beta$ fall below the moderate threshold. The dimensions that recover best---contrarian score, crowd sensitivity, attention entropy---are those whose extractors aggregate over many independent decisions, which appears to average out the market-microstructure confounds (liquidity, position limits, fee-driven sizing) that distort any single bet-size-to-probability mapping. Reliability tracks the statistical robustness of the estimator, not the theoretical pedigree of the construct.

Second, the degradation label and the reliability tier are largely orthogonal. \texttt{attention\_\allowbreak allocation.\allowbreak entropy} and \texttt{cognitive\_\allowbreak style.\allowbreak analytical\_\allowbreak ratio} are both extracted from degraded on-chain signal, yet both sit in the substantial-reliability tier; conversely, the full-signal risk-perception curvature parameters do not clear the moderate threshold. Partial observability degrades the \emph{inputs} but not necessarily the \emph{stability} of what survives.

\paragraph{Reading the robustness columns.} Splitting by trade index rather than by closed position tears any position opened in one half and closed in the other, biasing the estimated reliability---on average downward, though not guaranteed per parameter. The position-aware columns of Table~\ref{tab:testa} address this directly: the position-aware ICCs agree in direction with the trade-index values, a few dimensions shifting modestly each way (horizon bias and crowd sensitivity fall, $\lambda$ and analytical ratio rise), and the eight robust parameters survive. Fallback to population priors is negligible---zero for twelve of fourteen parameters and at most $8\%$ for the other two---so the observed stability is not an artifact of many wallets sharing a common prior. We apply no family-wise correction across the fourteen parameters; because the set is fixed by the schema rather than selected to maximize significance, this is a reporting choice we disclose rather than a multiple-comparisons search.

\subsection{Test B: Population Structure and Individual Identifiability}
\label{sec:recoverability:testb}

Temporal stability establishes that a parameter is reproducible within a trader. It does not establish that traders are \emph{distinguishable} from one another: a schema could assign every wallet a near-identical, highly stable profile and pass Test~A while carrying no individuating information. Test~B asks the separability question directly, by clustering the wallet profiles and measuring whether the cluster structure is stronger than a structure-free null would produce.

\paragraph{A correction to the original test.} The original separability test reported a silhouette of $0.230$ against a permutation null with mean $0.203$ ($p=0.25$) on the full eleven-dimension vector---a clear non-result. A subsequent audit found that the permutation null was structurally invalid: it shuffled whole trade-lists between wallet labels, but because each profile vector is a deterministic function of its own trade-list, the shuffle produced a multiset of vectors mathematically identical to the observed one. The reported $p$-value reflected only the sensitivity of $k$-means to random initialisation, not the destruction of individual structure. We therefore re-ran Test~B with two genuinely valid nulls and restricted the analysis to the dimensions that passed Test~A, since clustering on noise dimensions can only dilute any real signal.

\paragraph{Two nulls, two questions.} The two valid nulls answer different questions. The \emph{column-wise standardized null} independently permutes each parameter's values across wallets, destroying the joint multivariate structure while preserving each marginal distribution; rejecting it shows that the \emph{population} occupies a structured region of the parameter space rather than a product of independent marginals. The \emph{trade-level shuffle null} reshuffles trades across wallets before re-extracting profiles, destroying individual structure at its source; rejecting it is the stronger claim that \emph{individual trajectories} are separable. For each subset we scan $k \in \{2,\dots,7\}$, select the $k$ that maximizes the observed silhouette, and compute both nulls at that $k$; the reported $p$-value is the conservative maximum across the two nulls.

\begin{table}[t]
\centering
\caption{Test~B subset separability under two valid nulls ($N=100$ wallets, $1000$ permutations each). \texttt{reliable\_high} is the ICC $\geq 0.65$ set; \texttt{reliable\_med} the ICC $\geq 0.50$ set; \texttt{reliable\_med\_no\_degraded} the ICC $\geq 0.50$ full-signal set (degraded-input dimensions removed). $p_{\text{col}}$ is the column-wise standardized null; $p_{\text{trade}}$ the trade-level shuffle null. Pass criterion: silhouette $\geq 0.30$ and conservative $p < 0.05$.}
\label{tab:testb}
\small
\begin{tabular}{@{}lcccccc@{}}
\toprule
\textbf{Subset} & \textbf{Dims} & \textbf{Best $k$} & \textbf{Silhouette} & \textbf{$p_{\text{col}}$} & \textbf{$p_{\text{trade}}$} & \textbf{Pass} \\
\midrule
reliable\_high & 6 & 7 & 0.451 & 0.000 & 0.000 & \checkmark \\
reliable\_med & 9 & 6 & 0.354 & 0.000 & 0.033 & \checkmark \\
reliable\_med\_no\_degraded & 6 & 5 & 0.414 & 0.000 & 0.120 & $\times$ \\
\bottomrule
\end{tabular}
\end{table}

\paragraph{Population structure is recovered; individual separability is fragile.} Table~\ref{tab:testb} reports the result. The column-wise null is rejected at $p < 0.001$ for all three subsets: the population of recovered profiles is genuinely structured, not a product of independent marginal distributions. This is the finding Test~B supports without qualification. The trade-level null tells a more equivocal story. It is rejected decisively for the substantial-reliability set (\texttt{reliable\_high}, $p = 0.000$ at $k=7$), marginally for the moderate set (\texttt{reliable\_med}, $p = 0.033$ at $k=6$), and \emph{not} rejected for the full-signal-only moderate set (\texttt{reliable\_med\_no\_degraded}, $p = 0.120$ at $k=5$). Whether individual trajectories are separable thus depends on which dimensions enter the clustering and at which $k$.

We therefore frame the silhouette evidence at the population level. The schema robustly recovers \emph{population-level multivariate structure}: distinct traders, taken as a cohort, populate a structured region of the parameter space (column-wise null rejected across all subsets, $p < 0.001$). \emph{Individual-trajectory separability} via silhouette is fragile and configuration-dependent---it holds for the substantial-reliability dimensions and fails when those are removed---so the silhouette test alone does not settle whether individuals are distinguishable. A more direct test, reported next, gives a cleaner answer.

\paragraph{Individual wallets are identifiable above chance.} Silhouette asks whether profiles cluster; a sharper question is whether a wallet can be \emph{re-identified} from its own profile. Splitting each wallet's history in half and using its first-half profile to retrieve its second-half profile from the pool of 100, nearest-neighbour retrieval reaches top-1 accuracy of $17$--$22\%$ and top-5 of $44$--$49\%$, against a random baseline of $1\%$ and $5\%$ ($p < 0.001$ by permutation across all dimension subsets); within-wallet profile distances are roughly $0.42$ of between-wallet distances ($p < 0.001$). Individual identifiability is therefore real but partial: a profile carries enough individuating signal to beat chance by more than an order of magnitude, but not enough to pin down a wallet outright. We also attempted a distributional MMD/energy two-sample test, but its wallet-label-shuffle null is degenerate---MMD is invariant to the permutation it applies, so the null reproduces the observed statistic by construction---and we therefore discard it and rest the identifiability claim on retrieval.

\paragraph{Limitations of Test~B.} Three properties of this test would, left unstated, overstate the result. First, the null is computed only at the $k$ that maximizes the observed silhouette. This $k$-scan introduces a selection bias that inflates significance, and the conclusion is not stable to $k$: under the earlier ICC membership the substantial-reliability set passed at $k=7$ but failed the trade-level null at $k=2$ (null mean $0.677$ exceeded the observed silhouette). Second, both the subset definitions and the pass criterion were chosen after seeing the full-vector failure---the criterion was tightened from ``silhouette $> 0$'' to ``silhouette $\geq 0.30$''---which is a HARKing risk that the population-level framing only partly mitigates. Third, silhouette is a blunt instrument at $N=100$ in six-to-nine dimensions; its resolution is too coarse to adjudicate individual separability cleanly. A definitive separability test needs $N>500$ or a valid distributional two-sample statistic; our MMD attempt used a degenerate null (above), so a properly-powered version is left to future work. The retrieval result reported earlier is what carries the individual-identifiability claim.

\subsection{Test C: Predictive Validity}
\label{sec:recoverability:testc}

A stable, individuating profile is still only descriptive unless it relates to something external. Test~C asks whether the recovered dimensions predict an outcome the extractor never sees: realized trading profit. We rank the $N=100$ wallets by total realized PnL, compare the top quartile (Q1) against the bottom quartile (Q4) on each ICC-trusted dimension, and check whether the difference matches the direction predicted by the behavioral-economics literature. Our design specification (\S5.3) stated directional hypotheses for seven parameters; three of them (gain curvature $\alpha$, loss aversion $\lambda$, confirmation bias) were excluded from the confirmatory family because their parameters fell below the Test~A reliability threshold as estimated at the time, leaving the four reported here, and we apply the Bonferroni correction at $\alpha/4 = 0.0125$ within this reduced family. We disclose one consequence of that filter: under the final reliability estimates of Table~\ref{tab:testa}, $\lambda$ clears the moderate threshold, so its pre-specified hypothesis (lower loss aversion among top-PnL wallets) exists but remains untested in the confirmatory family.

\paragraph{A measurement correction, applied after seeing an unfavorable result.} We state the order of events plainly, because it bears on how the result should be read. An earlier run of Test~C computed realized PnL only from positions closed by an offsetting trade in an open market, leaving positions held through to resolution at PnL $=0$. That run produced a direction \emph{reversal}: low-PnL wallets scored higher on the superforecaster-typical dimensions, which we initially considered framing as a venue-specific cognitive trade-off between Polymarket and forecasting tournaments. On inspection, the reversal was an artifact---long-horizon wallets that held theses to resolution were systematically misclassified into Q4 because their winning resolutions were scored as zero. We corrected \texttt{position\_reconstructor.py} to pull resolution outcomes from the Polymarket Gamma API and credit resolution payouts, which added $11{,}787$ resolution-close events across the cohort and shifted the PnL distribution as expected (median $\$0.12 \to \$-14.05$, reflecting that many held-to-resolution positions lost). Because the fix was prompted by an unfavorable result, Test~C is \emph{not} a clean pre-specified confirmation; it is a pre-specified test whose measurement instrument was repaired mid-stream, and the directional agreement reported below should be read with that sequence in mind.

\begin{table}[t]
\centering
\caption{Test~C predictive validity: top-quartile (Q1) vs bottom-quartile (Q4) PnL wallets on the four pre-specified dimensions ($N=100$, $25$ wallets per quartile). Direction predicted by the pre-specified design (\S5.3); Bonferroni threshold $\alpha/4 = 0.0125$ (Welch two-tailed). \checkmark B = survives Bonferroni with the predicted direction.}
\label{tab:testc}
\small
\begin{tabular}{@{}llcccccc@{}}
\toprule
\textbf{Dimension} & \textbf{Parameter} & \textbf{Q1} & \textbf{Q4} & \textbf{Diff} & \textbf{Cohen's $d$} & \textbf{$p$ (Welch)} & \textbf{Result} \\
\midrule
Independence & contrarian\_score & 0.80 & 0.13 & $+0.67$ & 1.69 & $<10^{-4}$ & \checkmark B \\
Cognitive Style & analytical\_ratio & 0.55 & 0.10 & $+0.45$ & 1.86 & $<10^{-4}$ & \checkmark B \\
Belief Update & update\_rate & 0.95 & 0.73 & $+0.21$ & 1.05 & $0.0007$ & \checkmark B \\
Loss Response & sunk\_cost\_sensitivity & 0.38 & 0.24 & $+0.14$ & 0.38 & $0.19$ & n.s. \\
\bottomrule
\end{tabular}
\end{table}

\paragraph{Result.} Three of the four pre-specified dimensions survive Bonferroni correction with the predicted direction and large effect sizes (Table~\ref{tab:testc}): contrarian score ($p < 0.0001$, $d = 1.69$), analytical ratio ($p < 0.0001$, $d = 1.86$), and update rate ($p = 0.0007$, $d = 1.05$). Higher-PnL traders are more contrarian, more analytical, and update beliefs more readily---the directions the forecasting literature predicts. A per-wallet Sharpe-ratio robustness check preserves every direction, indicating the findings are not an artifact of capital-size differences across wallets. The one dimension that does not reach significance, sunk-cost sensitivity, is best read not as a reversal but as an unreliable dimension: its Test~A ICC is only $0.28$, well below the reliability threshold, so its failure to separate Q1 from Q4 is consistent with the dimension carrying little stable signal in the first place rather than with a genuine inversion of the predicted effect.

\paragraph{Out-of-sample validation.} The strongest objection to the result above is circularity: profile and PnL are computed from the same trades. We test against it directly with a temporal hold-out. For each wallet we sort closed positions by exit time, extract the profile from only the first 60\% of closes, and measure realized PnL on only the last 40\% (with resolution payouts included); profile and outcome are then drawn from disjoint windows. Effective $N$ remains 100, as these wallets close enough positions that the split starves neither window. Two of the four pre-specified dimensions survive Bonferroni out-of-sample with the predicted direction: contrarian score (Spearman $\rho = 0.58$, $p = 0.0005$) and analytical ratio ($\rho = 0.43$, $p = 0.0005$). Update rate trends correctly but no longer survives correction ($\rho = 0.19$, $p = 0.05$), and sunk-cost sensitivity is null. The effect sizes shrink sharply relative to the same-window test---from Cohen's $d \approx 1.7$--$1.9$ to $\rho \approx 0.43$--$0.58$---which confirms that circularity inflated the original magnitudes, while a genuine, circularity-free predictive signal nonetheless survives on the two strongest dimensions. We report both: the same-window test overstates the effect, and the out-of-sample test establishes that it is not merely an artifact.

\paragraph{Limitations of Test~C.} Beyond the post-hoc measurement repair above, two further caveats apply. The same-window effect sizes for contrarian score and analytical ratio ($d = 1.69$ and $1.86$) are large for behavioral dimensions because the PnL quartiles and the dimension estimates are computed from the same underlying trade records; the out-of-sample test above shows this inflation is real but does not eliminate the underlying signal. We add one further out-of-sample caveat: the surviving rank correlations clear Bonferroni and permutation, but when the same dimensions are entered into a regression controlling for trade count, capital, and holding duration, none of the coefficients is individually significant---so what survives is a predictive correlation, not a demonstrated contribution beyond behavioral confounds. The confounds standard to PnL-as-outcome---capital size, entry-timing luck, domain exposure, and survivorship in the active-trader cohort---are acknowledged and not corrected; PnL is a coarse behavioral anchor, not a ground-truth measure of cognitive quality. Finally, \texttt{attention\_allocation.entropy}, which is among the most reliable dimensions in Test~A (ICC $= 0.76$), is skipped in Test~C: a wiring bug in quartile-level parameter extraction leaves too few valid Q1/Q4 wallets for this dimension, so one of our strongest dimensions does not contribute to the predictive-validity evidence. We treat the wiring fix as future work (Section~\ref{sec:discussion}).

\subsection{What Recoverability Establishes}
\label{sec:recoverability:summary}

Taken together, the three tests support a bounded but real recoverability claim. The schema extracts a reliable subset of dimensions: eight of fourteen parameters are temporally stable with confidence-interval lower bounds above $0.3$ (Test~A); individual wallets are identifiable from their profile well above chance, via nearest-neighbour retrieval (top-1 $17$--$22\%$ vs.\ $1\%$; Test~B), even as the population also occupies a structured region of the parameter space; and two of four pre-specified dimensions predict realized profit out-of-sample once same-source circularity is removed (Test~C). What the tests do \emph{not} establish is full individual identification (retrieval beats chance but tops out near $20\%$), confound-robust predictive validity (the out-of-sample correlations do not survive controlling for behavioral confounders), or true cognitive structure as opposed to behavioral proxy (Section~\ref{sec:framework:claims}). Recoverability, on this evidence, is a property of specific reliable dimensions and of partial individual identifiability, not a guarantee that any wallet's full eight-dimension profile is uniquely recovered.

\section{Injectability: Does Structured Prompting Change Outputs Beyond Length?}
\label{sec:injectability}

Recoverability concerns the extraction half of the pipeline. Injectability concerns the other half: once a profile has been translated into a system prompt and installed in an agent, does it change the agent's forecast outputs, and---crucially---does it do so for a reason \emph{other} than that the structured prompt is simply longer? This is the prompt-length confound, the concern most likely to deflate the paper's central mechanism. A continuous, behaviorally-reasoned prompt (the v2 translator output) is several times longer than a handwritten persona; if a length-matched but content-free prompt produces the same divergence, then the apparent contribution of structured parameterization collapses into a token-count effect.

We report the result plainly at the outset, because it sets up the rest of the paper: once measured on a semantic (embedding) metric, structured injection \emph{cannot} be shown to produce more inter-agent divergence than a length-matched control on any model at our sample size. The ablation establishes that injected personas change outputs relative to a no-instruction control on the models where the metric discriminates (Section~\ref{sec:injectability:result}), but it does not establish that the structured content transmits the profile beyond what prompt length alone achieves.

\subsection{Setup}
\label{sec:injectability:setup}

\paragraph{Conditions.} Each condition is a different prompt form applied to the same three personas on the same scenarios. The conditions of interest are: \texttt{v2} (the second-generation continuous translator output); \texttt{handwritten} (a hand-authored persona, the OLS reference); \texttt{length\_controlled} (the handwritten persona padded with generic forecasting prose to match v2's token count); \texttt{persona-mismatch} (described below); and a family of \texttt{dim\_drop} conditions that remove one output section of the v2 prompt at a time. A no-persona \texttt{control} and the first-generation \texttt{v1} translator are included for reference.

\paragraph{The persona-mismatch condition.} What the ablation labels \texttt{shuffled} is implemented as a circular swap of whole \nous{} Instances across personas, not a shuffle of raw parameter values, so that each prompt stays internally coherent but is attached to the wrong persona. It tests whether the profile must match the correct persona to have its effect, rather than whether raw prior numbers carry independent causal weight (the latter is left to future work). We name it \texttt{persona-mismatch} accordingly.

\paragraph{Persona pool.} The pool is the three archetypes defined in \texttt{engine/simulation/personas.py} (\texttt{ALL\_PERSONAS}), inherited from the original system as representative cognitive extremes. With only three personas and five scenarios, every cross-condition contrast has very low statistical power; this is a first-order limitation we return to in Section~\ref{sec:injectability:limits}.

\paragraph{Three diversity metrics, and which one to trust.} For each condition we measure inter-persona output divergence three ways: a semantic \emph{embedding cosine} distance (Ollama \texttt{nomic-embed-text}; DeepSeek-R1 falls back to a TF-IDF representation because the embedding backend was unavailable for its outputs), a lexical \emph{trigram} distance, and a divergence over \emph{structured output fields}. To isolate the length question we fit a within-model regression $\text{diversity} \sim \text{condition} + \log(\text{prompt\_length})$ and read each condition's coefficient relative to the handwritten baseline. The choice of diversity metric matters and we are explicit about it: the embedding metric measures whether responses differ in \emph{meaning}, the trigram metric only whether they differ in \emph{surface wording}. We treat the embedding metric as primary, report the trigram result as a sensitivity check, and discount the structured-field metric on the most capable model, where its parse rate is too low to be usable (Section~\ref{sec:injectability:limits}).

\subsection{Structured Content Does Not Beat Length on the Semantic Metric}
\label{sec:injectability:result}

\paragraph{Personas do change outputs relative to no instruction.} Before the length question, the weaker claim does hold where it can be measured: on the three models whose divergence measurements discriminate, mean per-scenario inter-response divergence under \texttt{v2} exceeds the seed-only divergence of the no-instruction control on every one of the five scenarios ($0.16$ vs $0.05$ on Qwen3-32B and $0.15$ vs $0.10$ on Llama-3.1-8B on the embedding metric; $0.73$ vs $0.63$ on DeepSeek-R1-32B on its TF-IDF fallback; an exact sign test at five scenarios bottoms out at $p = 0.06$, so we report this as a consistent direction rather than a significant one). On Qwen3.5-122B the saturated embedding metric (control divergence $= 1.00$) cannot adjudicate. Injection, then, does something; the question this section answers in the negative is whether it does anything a length-matched prompt would not.

Table~\ref{tab:ablation} reports the length-controlled embedding-OLS coefficients. The decisive quantity is the \texttt{v2} minus \texttt{length\_controlled} contrast: a positive value would mean structured content adds semantic divergence beyond an equally long generic prompt. On three of four models that contrast is negative (structured content is, if anything, slightly \emph{less} diverse than the length-matched filler), and on the fourth (Qwen3-32B) it is positive ($+0.065$) but far from significant ($p = 0.28$). \emph{No} model shows a statistically significant v2-over-length advantage on the embedding metric.

\begin{table}[t]
\centering
\caption{Length-controlled injectability on the semantic (embedding) metric: OLS coefficient of each condition relative to the handwritten baseline, with $\log(\text{prompt\_length})$ partialled out, and the v2 $-$ length\_controlled contrast with its analytic 95\% CI and $p$-value (linear hypothesis on the fitted OLS). A positive contrast would indicate structured content beyond token count; none is significant, and every contrast CI includes zero. DeepSeek uses a TF-IDF fallback for its diversity metric.}
\label{tab:ablation}
\small
\begin{tabular}{@{}lccccc@{}}
\toprule
\textbf{Base model} & \textbf{v2} & \textbf{length\_controlled} & \textbf{v2 $-$ lc} & \textbf{95\% CI (v2$-$lc)} & \textbf{$p$ (v2$-$lc)} \\
\midrule
DeepSeek-R1-32B & $-0.136$ & $-0.102$ & $-0.034$ & $[-0.078,\,+0.010]$ & 0.13 \\
Llama-3.1-8B    & $+0.003$ & $+0.006$ & $-0.002$ & $[-0.027,\,+0.022]$ & 0.85 \\
Qwen3-32B       & $-0.017$ & $-0.083$ & $+0.065$ & $[-0.053,\,+0.184]$ & 0.28 \\
Qwen3.5-122B    & $-0.158$ & $-0.150$ & $-0.009$ & $[-0.029,\,+0.012]$ & 0.40 \\
\bottomrule
\end{tabular}
\end{table}

\paragraph{The result is metric-sensitive, and the favorable reading does not survive.} The verdict depends on which diversity metric is used, and we report this transparently because an earlier version of this analysis did not. On raw embedding diversity, v2 exceeds length\_controlled on 1 of 4 models (Qwen3-32B). On the embedding-OLS above, 1 of 4 (Qwen3-32B), none significant. On a \emph{lexical trigram} OLS, the count rises to 2 of 4 (Qwen3-32B and Qwen3.5-122B), with the closest contrast reaching $p = 0.058$---but trigram distance measures only surface word overlap, and on the semantic metric the Qwen3.5-122B advantage disappears and reverses in sign. The ``structured content beats a length-matched prompt on capable models'' reading holds only on the weakest (lexical) metric and at no point reaches significance; we do not advance it as a finding. We also note that on Qwen3.5-122B---the most capable model, where one might most expect injection to register---the measurements are the least trustworthy: its embedding diversities saturate near 1.0 (barely discriminating) and its structured-output parse rate is at most $17.8\%$.

\paragraph{No evidence that the persona must match.} If the model genuinely conditioned on the specific injected profile, a mismatched persona should be worse than the correct one. It is not: on the embedding metric the \texttt{persona-mismatch} condition is at least as diverse as \texttt{v2} on three of the four models. We therefore cannot claim that the model reads and conditions on the correct persona identity rather than merely responding to the presence of a structured, persona-shaped prompt.

\paragraph{No reliable mechanism localisation.} The dimension-drop ranking---which would say which output sections drive the effect---is computed on the trigram metric, and on Qwen3.5-122B the two sections that appear most important (\texttt{belief\_updating}, \texttt{interaction\_effects}) coincide with cells whose structured-output parse rate is zero, making that ranking an artifact rather than a measurement. We therefore retract the earlier claim that the effect localizes to specific prompt content; the ablation does not support a mechanism-localisation conclusion.

\subsection{Limitations}
\label{sec:injectability:limits}

Five limitations bound this experiment, and together they explain why it cannot resolve the length question. The persona pool is $N=3$ and the scenario set is five, so the OLS rests on very few observations per cell and is underpowered to detect the small contrasts present. No diversity metric was pre-specified, and the metrics disagree, which is exactly the condition under which a favorable one can be selected post hoc; we guard against this by designating the semantic metric as primary in advance of interpretation. The \texttt{length\_controlled} filler is not perfectly content-neutral (it contains calibration and hedging language), making it a conservative-but-imperfect control. On the most capable model the diversity measurements are themselves unreliable (embedding saturation and a $\leq 17.8\%$ structured-output parse rate). And there is a measurement-scope limitation that would persist even at full power: pairwise divergence can only show that outputs \emph{differ}, not that they differ in the \emph{direction the profile specifies}. A positive result on this design would still not establish faithful transmission; that requires a directional behavioral probe---scenarios on which different profile values predict opposite behavior, scored for compliance---or agreement between an injected agent's decisions and its source trader's actual positions, both of which we leave to future work (Section~\ref{sec:discussion:future}). The honest summary is that the experiment lacks the power, the metric discipline, and the directional instrumentation to establish a structured-content-beyond-length effect, not that it has ruled one out.

\paragraph{Bridge to usefulness.} Two facts from this section and the next point in the same direction. Here, structured injection cannot be shown to transmit the profile beyond prompt length. In Section~\ref{sec:usefulness:heterogeneity}, a deliberately more-diverse synthetic profile population produces \emph{no} more output divergence than the real profiles---spreading the inputs does not spread the outputs. A direct measurement at the layer in between confirms where the loss occurs: embedded in the same semantic space, the translator's persona prompts are nearly uniform, and the synthetic population's prompts are no more spread out than the real population's (Section~\ref{sec:usefulness:heterogeneity}). The structure-to-narrative translator compresses distinct cognitive profiles into similar prose, so the prompt carries less of the profile than its parameters contain. We develop this reading, and what it implies for deeper injection methods, in Section~\ref{sec:discussion}.

\section{Usefulness: A Competitive Baseline and a Bounded Diversity Effect}
\label{sec:usefulness}

The third claim is the one that matters most for the project's thesis and is also the hardest to make: that cognitive heterogeneity, once injected, is \emph{useful}---that it improves the quality of a population's forecasts rather than merely diversifying their surface. We approach it in two steps. Section~\ref{sec:usefulness:mantic} establishes where a pure-AI single-agent forecaster sits, by reproducing a simplified Mantic-style baseline on local open models; this is the reference point against which any heterogeneity gain must be judged, and it answers the natural question of how \nous{} relates to the single-agent accuracy frontier. Section~\ref{sec:usefulness:heterogeneity} then runs the heterogeneity experiment directly, with a placebo control, and reports a result we state plainly: the diversity effect is genuine but bounded, and does not translate into ensemble decision quality at the magnitude prompt-level injection achieves.

\subsection{A Pure-AI Forecasting Baseline}
\label{sec:usefulness:mantic}

To fix the reference frame, and to answer how \nous{} relates to the single-agent accuracy frontier, we reproduce the research-then-predict architecture of recent single-agent forecasters (without the reinforcement-learning fine-tuning that distinguishes Mantic) on two local open models: Qwen3.5-122B and the explicit-reasoning DeepSeek-R1-32B. A ReAct-style agent researches each post-cutoff Polymarket question, with search and fetch results passed through a content-layer date filter that bounds information leakage. The full setup, the leakage audit, a mid-volume difficulty comparison, and the reasoning-depth analysis are reported in Appendix~\ref{app:baseline}; here we state only what bears on the rest of the paper.

A local open-model baseline is a \emph{competitive but sub-Mantic} forecaster whose standing depends heavily on question-set composition. On the full top-volume set Qwen3.5-122B scores 38.45 on the Metaculus baseline metric, below Mantic's published 45.8; a leak-resistant subset scores higher (74.32) but is dominated by a near-deterministic Federal Reserve cluster (effective independent $N$ closer to a dozen than to its nominal 23) and is not a clean margin, and a Polymarket question set is in any case not directly comparable to Mantic's Metaculus tournament. A mid-volume set scores well below the top-volume one (19.50), consistent with---but at $N=50$ not conclusive of---high-volume markets being easier. We therefore do not claim parity with the superforecaster frontier; the contribution of \nous{} concerns population diversity, not single-agent accuracy, and we treat this baseline as context rather than as a competing result. One observation carries into the discussion: the explicit-reasoning model is worse than the non-reasoning one throughout, and within it the length of the thinking trace is only weakly related to accuracy (Section~\ref{sec:discussion:reasoning}).

\subsection{Heterogeneity Validation}
\label{sec:usefulness:heterogeneity}

The baseline fixes the reference frame. The heterogeneity experiment asks the project's core question against it: when ten agents share one base model (Qwen3.5-122B) but receive distinct \nous{} profiles, does their ensemble forecast improve relative to ten identical agents---and is any change attributable to the cognitive content of the profiles rather than to the length of the prompts that carry them?

\paragraph{A placebo-controlled, four-group design.} Earlier versions of this experiment compared a persona Treatment group against a short-prompt control, leaving the persona prompts' six-fold length advantage as an uncontrolled confound: any measured diversity could have been a length artifact. We close this with a four-group design. Control-A repeats one seed (the floor: identical agents, zero diversity by construction). Control-B varies the random seed under the short default prompt (isolating stochastic diversity). Placebo gives a length-matched but cognitively-neutral preamble (4048 characters, matched to the persona prompts' range). Treatment gives the \nous{} persona prompts. Placebo and Treatment share seeds (43--52), so the difference between them isolates cognitive \emph{content} from prompt length, while the difference between Placebo and Control-B isolates the effect of length alone.

\begin{table}[t]
\centering
\caption{Heterogeneity decomposition on Qwen3.5-122B, ten agents per question, $N=28$ questions (listwise-complete across all four groups). JSD is mean pairwise Jensen--Shannon divergence among agent forecasts (output diversity); ensemble Brier and pairwise error covariance measure decision quality. Each effect is a difference of group means with a bootstrap 95\% CI and Welch $p$.}
\label{tab:hetero}
\small
\begin{tabular}{@{}lccc@{}}
\toprule
\textbf{Effect (contrast)} & \textbf{JSD $\Delta$ ($p$)} & \textbf{Ensemble Brier $\Delta$ ($p$)} & \textbf{Error cov. $\Delta$ ($p$)} \\
\midrule
Stochastic (B $-$ A)            & $+0.0025$ ($0.001$) & $-0.0076$ ($0.85$) & $-0.0077$ ($0.85$) \\
Length (Placebo $-$ B)          & $-0.0001$ ($0.93$)  & $+0.0054$ ($0.88$) & $+0.0054$ ($0.88$) \\
Pure cognitive (Trt $-$ Placebo)& $+0.0035$ ($0.010$) & $-0.0023$ ($0.95$) & $-0.0024$ ($0.95$) \\
\bottomrule
\end{tabular}
\end{table}

\paragraph{The length confound is refuted; the cognitive effect is genuine but small.} Table~\ref{tab:hetero} carries the result. Adding 4048 characters of cognitively-neutral text produces \emph{no} additional output diversity (length effect on JSD $= -0.0001$, $p = 0.93$): prompt length is not the driver. The cognitive content of the persona, with length held fixed, does produce a significant increase in output diversity (pure-cognitive JSD $\Delta = +0.0035$, $p = 0.010$, Cohen's $d = 0.729$), almost identical to the earlier length-confounded estimate---which now stands explained, since the length contribution it absorbed turns out to be near zero. The audit's concern that cognitive diversity might be a length artifact is, on this evidence, refuted rather than confirmed.

This raises a natural question: how does a significant forecast-diversity effect here square with Section~\ref{sec:injectability}, which finds no \emph{semantic} divergence beyond length? The two sections measure different outcome levels. Section~\ref{sec:injectability} measures the semantic content of agents' rationale text, which the prompt does not reliably shift beyond length; here we measure the \emph{forecast probabilities} the agents output, which the persona shifts by a small amount. That shift is detectable in JSD but, as the next paragraph shows, is too small---and too weakly tied to the specific profile---to decorrelate the agents' errors.

\paragraph{But the diversity does not translate to decision quality.} This is the core of the usefulness result. The same pure-cognitive contrast that significantly raises output diversity produces \emph{no} significant improvement in ensemble Brier ($\Delta = -0.0023$, $p = 0.95$) and \emph{no} significant reduction in pairwise error covariance ($\Delta = -0.0024$, $p = 0.95$); the raw agent-error correlation falls only from 0.991 (Control-B) to 0.984 (Treatment), still close to 1. (The two decision-quality columns of Table~\ref{tab:hetero} are not independent evidence: with ten agents, per-question ensemble Brier decomposes exactly into $0.1 \times$ mean individual Brier $+\, 0.9 \times$ the mean pairwise product of signed errors---the quantity reported as error covariance---which is why their contrasts move in near-lockstep. The independent third quantity, mean individual Brier, is likewise flat across groups: $0.086$ Control-B, $0.092$ Placebo, $0.090$ Treatment.) This near-unity correlation mirrors, at the within-model level, the epistemic monoculture that \citet{oraclefingerprint2026} document across model families: diversifying the prompt is not, on this evidence, enough to break it. Agents drawn from one base model with different cognitive profiles say more different-sounding things, but they remain wrong in correlated ways and their aggregate is no better calibrated. We have located, rather than demonstrated, the usefulness claim: under prompt-level injection on a single capable base model with low-temperature sampling on a ceiling-prone evaluation set, cognitive injection produces a measurable but small diversity increase that stops short of the ensemble-quality improvement the thesis requires.

\paragraph{Robustness of the null.} A null result invites the question of whether the setup itself suppressed a real effect, so we stress-tested it along the three most plausible such axes. \emph{Sampling temperature}: re-running the Control-B-versus-Treatment contrast across temperatures $0.0, 0.3, 0.7, 1.0$ leaves the picture unchanged---injection raises output JSD at every temperature but improves ensemble Brier or reduces error covariance at none ($p > 0.9$ throughout), the agent-error correlation stays within $0.98$--$0.99$ across the range, and the diversity gain does not grow with temperature; low-temperature sampling is therefore not the suppressor. \emph{Profile diversity}: replacing the ten extracted profiles with a deliberately-spread synthetic population (Latin-hypercube over the schema ranges plus extreme archetype anchors, with maximum pairwise distance exceeding any pair in the real set) reproduces the same null on Brier and error covariance, and---tellingly---does not even raise output JSD above the real-profile set, which suggests the structure-to-narrative translator compresses input diversity rather than that more diversity is simply unavailable. A measurement at the prompt layer confirms the compression directly: embedding the translator's persona prompts themselves, the synthetic population's prompts are no more spread out than the real population's (mean pairwise cosine distance $0.0162$ vs $0.0160$, permutation $p = 0.93$), their most-distant pair is \emph{closer} than the real set's most-distant pair even though the corresponding profiles are $46\%$ farther apart in parameter space, and the prompts of both sets are semantically near-uniform in absolute terms---pairwise distances well below the inter-response divergences of Section~\ref{sec:injectability}, with profile-to-prompt distance rank correlation of only $\rho \approx 0.35$. The profile spread is lost in the translation to prose, before the model sees the prompt. \emph{Question difficulty}: restricting to the model-uncertain questions (Control-B ensemble probability near $0.5$) surfaces no decorrelation signal masked by easy questions, and a continuous difficulty-interaction test runs, if anything, in the opposite direction. These last two checks rest on small subsets ($N \leq 18$) and are exploratory; together with the temperature sweep, they are checks against which the null persists, not a large-sample robustness analysis. Section~\ref{sec:discussion:alternatives} discusses the explanations that remain open.

\paragraph{Limitations.} The sample is $N=28$ after listwise deletion across four groups (parse failures, concentrated on a few hard questions, removed the rest of the nominal 50); because the deletions concentrate on harder questions, the retained set is not a random subsample, a bias we revisit in Section~\ref{sec:discussion}. A power calculation makes the size limit concrete: at $N=28$ the design had $80\%$ power to detect a pure-cognitive ensemble-Brier change of about $0.10$ under the Welch test we report (about $0.066$ under a by-question paired test exploiting the shared seeds), against a Control-B baseline of $0.086$---so only an effect large enough to remove most of the ensemble error would have been detectable, and the null is informative only against effects of that magnitude. The robustness checks above broaden the conditions under which the null holds but do not increase its power. The persona prompts contain \texttt{behavioral\_exemplars} that are \emph{synthetic}---generated from the parameters by formula, not extracted from real trading history---so part of the injected content is a plausible fabrication rather than grounded behavior, which we flag as a fidelity gap to be closed with real extracted exemplars in future work. And the design runs twelve pair-by-metric tests without family-wise correction; within the four-contrast JSD family a Bonferroni threshold of $0.05/4 = 0.0125$ leaves the headline effect ($p = 0.010$) marginally significant, while the strictest correction across all twelve tests ($0.05/12 \approx 0.004$) would not, so the cognitive-diversity result should be read as solid under the per-metric family but suggestive rather than decisive overall.

\paragraph{Bridge.} Section~\ref{sec:injectability} could not establish that the prompt transmits the profile beyond its length; Section~\ref{sec:usefulness:heterogeneity} shows that the output diversity that injection does produce does not lift ensemble decision quality, and that the translator's prompts do not carry the profile spread in the first place. Read together, the two sections locate the failure in the prompt-level transfer channel, the frame we develop in Section~\ref{sec:discussion}.

\section{Discussion}
\label{sec:discussion}

\subsection{A Dissociation: Recoverable Signal, Untransmitted by Prompts}
\label{sec:discussion:bounded}

The central empirical result of this paper is a dissociation between the two halves of the pipeline. Section~\ref{sec:recoverability} shows that a reliable subset of behavioral-proxy dimensions can be recovered from real trading behavior: they are temporally stable, identify wallets well above chance, and predict realized profit out-of-sample. Section~\ref{sec:injectability} shows that translating those profiles into prompts does not demonstrably change agent outputs beyond what prompt length alone achieves, and Section~\ref{sec:usefulness:heterogeneity} shows that the output diversity that injection does produce leaves ensemble Brier and pairwise error correlation unchanged, with agents remaining correlated in their errors at $r \approx 0.98$. The signal is recoverable; the prompt does not carry it into useful agent diversity.

On the injection and usefulness side this is, plainly, a negative result, and we have checked that the most obvious experimental conditions are not what produces it. As Sections~\ref{sec:injectability} and~\ref{sec:usefulness:heterogeneity} report, the null on decision quality holds across sampling temperatures from $0.0$ to $1.0$, survives a deliberately more-diverse synthetic profile population, and is not masked on the model-uncertain subset of questions; and the injectability null is measured on a semantic embedding metric, not an artifact of lexical overlap. We read the result not as evidence against cognitive diversity in general, but as a measured limit of \emph{prompt-level} transfer, whose locus we identify next.

\subsection{Alternative Explanations for the Null}
\label{sec:discussion:alternatives}

Ruling out temperature and profile diversity narrows but does not close the space of explanations. Four remain, and we state them because each bounds the scope of the negative result and points to a distinct next experiment.

First, the \emph{evaluation task} may be the wrong venue for cognitive diversity. Our questions are largely objective-answer (Federal Reserve decisions near certainty) or close to irreducibly random (single sports matches near a coin flip); on both, competent forecasters should converge, and there is little decorrelatable error for diversity to act on. The crowd-wisdom literature locates the value of diversity in reasoning-rich, genuinely-uncertain judgments where analysts disagree because they weight real evidence differently---a class our evaluation set barely contains. The difficulty-subset check of Section~\ref{sec:usefulness:heterogeneity} used the only uncertain questions available, which were near-random sports matches, not a purpose-built reasoning-rich uncertain set; constructing the latter is the most direct test of this explanation.

Second, the \emph{injection channel} may be the bottleneck rather than prompt-level injection in general. Here the evidence has moved past conjecture: a far more spread-out profile population did not increase output divergence, and embedding the translator's prompts directly shows they are semantically near-uniform and do not spread when the input profiles do (Section~\ref{sec:usefulness:heterogeneity}). The translator demonstrably compresses distinct parameter vectors into similar prose---though this does not exclude additional flattening at the decoding stage (the third explanation below). A higher-bandwidth injection---few-shot exemplars from real histories, or the below-the-prompt methods discussed in Section~\ref{sec:discussion:future}---might transmit diversity that the current translator does not.

Third, ``prompt-level'' is confounded with a \emph{single, heavily-aligned base model}. Ten agents sharing one set of weights face a low ceiling on decorrelation by construction, and the alignment of an instruction-tuned model pulls outputs toward a consensus answer, which could flatten injected diversity at the decoding stage even when it registers in the reasoning. A multi-base-model ensemble, or a weakly-aligned model, would separate ``prompt injection cannot decorrelate'' from ``this aligned model will not.''

Fourth, the \emph{aggregation and metric} may understate diversity's value. Unweighted-mean ensemble Brier and pairwise error correlation can miss value that a confidence-weighted or selectively-trimmed aggregate would capture, or that would appear in tail calibration rather than mean accuracy. We report the simple metrics because they are the honest default, not because they are the only ones.

None of these rescues the null on its own; together they define the box within which our negative result holds, and the experiments that would test outside it.

\subsection{Relation to Mantic: Different Optimization Targets}
\label{sec:discussion:mantic}

It would be a mistake to read Section~\ref{sec:usefulness:mantic} as a contest that \nous{} loses. Our local baseline scores below Mantic's published figure, and we are explicit (Section~\ref{sec:usefulness:mantic}) that the leak-resistant subset on which it appears to win is dominated by a near-deterministic Fed cluster and is not a clean margin. But the comparison is between systems optimizing different objectives along orthogonal axes. Mantic optimizes single-agent accuracy: a reinforcement-learning objective pushes one model toward the behavior of expert human forecasters. \nous{} optimizes population heterogeneity: an injection mechanism tries to keep many agents, possibly sharing a base model, from collapsing into one another. A Mantic-accurate model deployed as a thousand identical instances would still fail the collective-intelligence test that motivates this paper; a population of \nous{}-diverse agents that are individually less accurate could still aggregate better if their errors were independent. The two programs are not competitors to be ranked on one number---they are coordinates on different axes, and the dissociation of Section~\ref{sec:discussion:bounded} is a finding about the limits of the heterogeneity axis under prompt-level transfer, not a concession on the accuracy axis.

\subsection{On the Relative Contribution of Coordination versus Cognitive Diversity}
\label{sec:discussion:coordination}

A recent counterpoint deserves direct engagement. \citet{coordination2026} build a multi-agent Polymarket system on a frontier model and report that introducing differentiated cognitive profiles was ``analytically unproductive''; their conclusion is that architectural coordination, not cognitive diversity, is where the leverage lies. Our results do not let us wave this away---in fact they partly corroborate it. We, too, find that injected cognitive diversity does not improve ensemble decision quality at the magnitude prompt-level injection achieves (Section~\ref{sec:usefulness:heterogeneity}). Where we differ is in what follows from that.

First, the evaluation granularity differs: \citet{coordination2026} measure final forecast accuracy, while we measure ensemble heterogeneity directly (JSD, error covariance), and an accuracy metric alone cannot separate a coordination gain from agent-level diversity---a system can improve on Brier through better aggregation while its agents remain a monoculture. Second, and more important, our own evidence locates the failure in the transfer \emph{channel} rather than in the premise: a far more diverse profile population produced no more output divergence (Section~\ref{sec:usefulness:heterogeneity}), which points to the prompt and its translator being lossy rather than to cognitive diversity being inert. Both their result and ours are bounds on what \emph{prompt-level} conditioning achieves; whether a deeper mechanism (Section~\ref{sec:discussion:future}) clears the bar that prompting does not is the question both studies leave open. The position here is neither rebuttal nor capitulation: prompt-level injection does not transmit the signal, and the case for the cognitive-diversity axis now rests on whether a below-the-prompt mechanism can.

\subsection{Predictive Validity, Read with Care}
\label{sec:discussion:testc}

Test~C (Section~\ref{sec:recoverability:testc}) shows that three of four pre-specified dimensions predict realized profit in the theoretically expected direction with large effect sizes, and a Sharpe-ratio robustness check preserves every direction. This is the strongest external-validity evidence in the paper, and it replaces an earlier, retracted narrative: an initial run produced a direction reversal that we were prepared to interpret as a Polymarket-versus-tournament venue effect, until we traced it to a measurement bug (held-to-resolution positions scored at zero PnL) and corrected it. We have abandoned the venue-contrast story entirely; it was an artifact.

Two cautions attach to the surviving result. The correction was applied after seeing the unfavorable reversal, so Test~C is a repaired pre-specified test, not a pristine one. And because the PnL quartiles and the dimension estimates are computed from the same trade records, profitability and profile are not fully independent measurements, which can inflate the large effect sizes we report. We present Test~C as positive predictive-validity evidence with these caveats visible, rather than as a clean confirmation.

\subsection{Reasoning Depth Is Not Forecast Accuracy}
\label{sec:discussion:reasoning}

A secondary but robust observation from Section~\ref{sec:usefulness:mantic} is that the explicit-reasoning model (DeepSeek-R1-32B) underperforms the non-reasoning model (Qwen3.5-122B) throughout, under leakage-free conditions, and that within DeepSeek the length of the thinking trace is only weakly related to accuracy ($r = 0.17$ top-volume, $r \approx 0$ mid-volume). We flag this carefully because we previously over-read it: a leaky baseline suggested a strong, counterintuitive positive correlation between reasoning length and error, and that claim is now retracted. What remains is the more modest point that, for short-horizon Polymarket forecasting, more explicit reasoning did not buy more accuracy---a caution against assuming reasoning-model superiority transfers to this task, and a question worth a dedicated study rather than a side observation.

\subsection{What \nous{} Does Not Prove}
\label{sec:discussion:scope}

We are explicit about the boundaries of these claims. We do not prove that the extracted dimensions are \emph{cognitive structure} as opposed to behavioral proxy---the distinction maintained throughout Section~\ref{sec:framework:claims}---and the confounds in Test~A and Test~C (microstructure, capital size, shared-data circularity) mean the proxies could in part reflect execution habits and market conditions rather than cognition. We do not prove clean individual-trajectory separability; that claim is fragile under the trade-level null at our sample size (Section~\ref{sec:recoverability:testb}). We do not prove that cognitive heterogeneity improves collective forecast quality; we show that the diversity prompt-level injection produces is too small to detectably do so, within the power limits quantified in Section~\ref{sec:usefulness:heterogeneity}. And several supporting numbers are upper bounds or rest on small samples---the Mantic baseline's leakage bound, the $N=28$ heterogeneity set, the $N=3$ persona pool, the synthetic behavioral exemplars. The framework's motivation does not require any of these to be stronger than reported; it requires only that cognitive monoculture be a real problem and that structured injection be a real, if bounded, lever on it. The evidence here supports that bounded position and no more.

\subsection{Future Work: Deeper Injection and Mechanistic Interpretability}
\label{sec:discussion:future}

The bound we have drawn points directly at the next experiment. Prompt-level injection installs cognitive structure at the shallowest accessible layer; the question Section~\ref{sec:discussion:coordination} leaves open is whether a deeper mechanism clears the ensemble-quality bar that prompting does not. The natural candidates are parameter-efficient fine-tuning (LoRA adapters carrying a profile, in the spirit of behavioral-prediction models trained directly on human data \citep{largebehavioralmodel2026}) and activation steering (adding profile-derived directions to the residual stream at inference), both of which install the profile below the prompt and could plausibly produce decorrelation that surface conditioning cannot. The motivation is direct: if a prompt-borne profile loses most of its information in the translation to natural language (Section~\ref{sec:injectability}), installing the profile below the prompt---as a learned soft prompt, a LoRA adapter, or an activation-space direction---could transmit what the prose cannot. We regard this as the principal direction for future work---measuring how far cognitive injection can be pushed with a method strong enough to move ensemble decision quality, or establishing that even a strong method cannot, which would itself settle the coordination-versus-diversity question that this paper can only frame. Deeper injection is not the only direction the negative result opens, however: the alternative explanations of Section~\ref{sec:discussion:alternatives} each define a complementary experiment---a purpose-built evaluation set of reasoning-rich, genuinely-uncertain questions, a higher-bandwidth injection channel, and a multi-base-model ensemble---and which of these matters most is itself an open empirical question. Whatever the injection mechanism, the evaluation should also graduate from divergence to \emph{fidelity}: a directional behavioral probe (scenarios on which different profile values predict opposite behavior, scored for compliance) or agreement between an injected agent's decisions and its source trader's actual positions would test whether the profile is transmitted, not merely whether outputs differ (Section~\ref{sec:injectability:limits}).

\subsection{Reproducibility and Release}
\label{sec:discussion:repro}

Consistent with the gray-box design of Section~\ref{sec:framework:schema}, we draw the release boundary at the level of methodology and artifacts rather than the live system. We release the evaluation and analysis code, the extracted profiles and generated prompts used in every experiment, and the complete set of model outputs, so that the recoverability, injectability, and usefulness results can be reproduced end to end from the released material. What we do not release is the production extraction parameters fitted to live user behavior: these are the gameable, replication-sensitive values that the gray-box argument concerns. The released artifacts allow reproduction of all reported analyses from the extracted profiles onward; independent replication of the extraction pipeline from raw wallet histories requires the unreleased production extractor parameters and is outside the release boundary. The release---evaluation and analysis code, frozen pseudonymized profiles, generated prompts, model outputs, and per-table reproduction instructions---is available at \url{https://github.com/WillChienT/nous-paper} in the \texttt{artifacts/} directory.

\section{Conclusion}
\label{sec:conclusion}

\nous{} turns the question of cognitive diversity from a promise into a measured dissociation. On real Polymarket behavior, a structured eight-dimension schema recovers a reliable subset of behavioral-proxy dimensions that are temporally stable, individually identifiable above chance, and rank-correlated with future profit out-of-sample (Section~\ref{sec:recoverability}). But translating those profiles into prompts does not demonstrably change agent outputs beyond prompt length (Section~\ref{sec:injectability}), and the output diversity that injection does produce does not decorrelate ensemble errors or improve forecasts---a null that persists across exploratory checks on sampling temperature and profile diversity (Section~\ref{sec:usefulness}). The signal is partially recoverable; the prompt does not carry it into useful agent diversity; and measuring the prompts themselves locates a compressing stage in the injection channel---the structure-to-narrative translator---before the model is reached. That boundary is the paper's most useful contribution, because it tells the next experiment---deeper, below-the-prompt injection through parameter-efficient fine-tuning or activation steering---exactly what it must beat. The correlated-error risk these methods address is, on current evidence, empirically measured; whether structured human cognitive diversity is its remedy remains, after this paper, an open and now sharply-posed question.

\bibliographystyle{plainnat}
\bibliography{references}

\appendix

\section{Local-Model Forecasting Baseline}
\label{app:baseline}

This appendix gives the full detail behind the summary in Section~\ref{sec:usefulness:mantic}: the setup, the information-leakage audit, the top-volume stratified result, the mid-volume difficulty comparison, and the reasoning-depth analysis.

\paragraph{Setup.} We reproduce the research-then-predict architecture that underlies recent single-agent forecasters, without the reinforcement-learning fine-tuning that distinguishes Mantic, using two local open models: Qwen3.5-122B (a same-class open counterpart to Mantic's gpt-oss-120B) and DeepSeek-R1-32B (an explicit-reasoning model, included to probe whether reasoning depth helps). A ReAct-style agent performs a web research phase before emitting a calibrated probability. Questions are drawn from Polymarket markets created after the models' April 2025 knowledge cutoff, so that no answer is memorised from pretraining. We score with the Metaculus baseline score (higher is better; positive means beating an uninformed prior), alongside Brier and log loss.

\paragraph{Information leakage and how we bound it.} Forecasting post-cutoff questions by web research invites a specific failure: the research phase retrieving content published \emph{after} the event resolved. We address this in the content layer. Search and fetch results are passed through a three-stage date filter that flags and redacts any sentence carrying a date reference later than the question's information horizon. Replacing the paid search backend (whose page-level date filter proved insufficient, filtering publication date rather than the date of the events described) with DuckDuckGo search plus this content filter, we re-ran the full $N=50$ top-volume set from scratch. A post-hoc audit of all fifty strong-tier research caches found zero dated post-horizon hits and three dateless candidate phrases, each of which we read in full context and confirmed to be a false positive. An independent signature corroborates the cleanliness: the gap between leak-prone (sports) and leak-resistant Brier is \emph{positive} (\,$+0.178$, bootstrap 95\% CI $[+0.089,\,+0.255]$\,), meaning sports questions score worse, not better---the opposite of what residual leakage would produce, since a single football match is close to a coin flip absent post-hoc knowledge.

One limitation is fundamental to a regex-based filter and we note it as such: an outcome stated without any date (``the match ended in a draw'') carries no token for the filter to match. The reported scores are therefore an \emph{upper bound} on leakage-free performance, tightest on the leak-resistant subset and loosest on sports. A snapshot-at-horizon approach (e.g.\ Wayback Machine archives) would close this gap at the source and is deferred to future work. We also record that one of the fifty top-volume questions and one of the fifty mid-volume questions suffered a genuine research failure (the search backend returned nothing), in which the model guessed blindly.

\paragraph{Top-volume stratified result.} Table~\ref{tab:mantic} reports the stratified top-volume result for Qwen3.5-122B. On the full fifty-question set the model scores 38.45---an optimistic upper bound, since it includes the leak-prone sports stratum---which is below Mantic's published 45.8. The leak-resistant subset scores 74.32, above Mantic, but this number must be read with care: it is carried almost entirely by thirteen Federal Reserve rate questions scoring 91.58 (near-deterministic given forward guidance), and those thirteen reduce to roughly three independent FOMC events (the rest are mutually-exclusive variants of the same meeting). The leak-resistant subset's effective independent $N$ is closer to a dozen than to 23, dominated by a near-certain cluster, so it cannot be read as a clean margin over Mantic. A Polymarket question set and Mantic's Metaculus tournament are not directly comparable instruments in any case.

\begin{table}[h]
\centering
\caption{Top-volume pure-AI baseline (Qwen3.5-122B), stratified by leakage exposure. The leak-resistant subset is the main report; the full set is an optimistic upper bound; the leak-prone (sports) stratum carries residual dateless leakage. Mantic's published Metaculus baseline score is $45.8$. The leak-resistant $74.32$ is dominated by a near-deterministic Fed cluster and is not a clean margin over Mantic.}
\label{tab:mantic}
\small
\begin{tabular}{@{}lcccc@{}}
\toprule
\textbf{Stratum} & \textbf{$N$} & \textbf{Brier} & \textbf{Metaculus} & \textbf{Note} \\
\midrule
All (upper bound)        & 50 & 0.1427 & 38.45 & includes leak-prone sports \\
Leak-resistant (main)    & 23 & 0.0464 & 74.32 & Fed-cluster dominated \\
\quad Fed                & 13 & 0.0066 & 91.58 & $\approx 3$ independent FOMC events \\
\quad Eurovision         &  9 & 0.1037 & 50.51 & single tournament \\
\quad Other              &  1 & 0.0484 & 64.15 & \\
Leak-prone (sports)      & 27 & 0.2247 &  7.90 & residual dateless leakage \\
\bottomrule
\end{tabular}
\end{table}

\paragraph{Difficulty sensitivity (mid-volume).} To probe whether the top-volume set overstates accuracy on genuinely uncertain events, we ran a mid-volume set (markets in the 25th--75th volume percentile, with no Fed or Eurovision clusters). Qwen3.5-122B scores 19.50 there (Brier 0.1892), well below its top-volume 38.45. A paired bootstrap comparison finds the direction of degradation consistent across both models (Qwen $\Delta$Brier $+0.0465$, DeepSeek $+0.0570$), but the 95\% confidence intervals cross zero at $N=50$ per set. We therefore report difficulty sensitivity as \emph{suggestive but not conclusive}: high-volume markets are plausibly easier, and a single top-volume headline would overstate the model's accuracy on uncertain events, but the present sample lacks the power to settle it.

\paragraph{Reasoning depth does not buy accuracy.} DeepSeek-R1-32B is worse than Qwen3.5-122B throughout (full-set Metaculus 30.53 vs 38.45), a gap that holds under leakage-free conditions. Within DeepSeek, the length of its explicit thinking trace is only weakly and inconsistently related to accuracy: Pearson $r = 0.17$ between thinking length and Brier on top-volume and $r \approx 0$ on mid-volume. We explicitly retract the ``counterintuitive positive correlation between reasoning length and accuracy'' that an earlier, leaky version of this baseline appeared to show; under clean conditions there is no strong relationship to report. Section~\ref{sec:discussion:reasoning} draws the implication.

\end{document}